\documentclass[11pt]{article}

\usepackage[final]{acl}

\usepackage{times}
\usepackage{latexsym}
\usepackage{booktabs}
\usepackage{makecell}
\usepackage{adjustbox}
\usepackage{array}
\usepackage{amssymb}
\usepackage{tcolorbox}

\tcbuselibrary{listings,breakable,skins}
\newcolumntype{C}[1]{>{\centering\arraybackslash}p{#1}}

\newcommand{\cmark}{\checkmark}
\newcommand{\xmark}{$\times$}
\usepackage[T1]{fontenc}
\usepackage[utf8]{inputenc}

\usepackage{microtype}

\usepackage{inconsolata}

\usepackage{graphicx}
\usepackage{amsmath} 
\title{S-MARC: Causal Streaming Reasoning for Full-Duplex Conversational Behavior Modeling}

\author{
\textbf{Dingkun Zhou}\textsuperscript{1,3,*,\textdagger}\thanks{%
\textsuperscript{*}Equal contribution.
\textsuperscript{\textdagger}Corresponding authors:
\texttt{jackkun818@gmail.com}, \texttt{jiachenlian@berkeley.edu}.},
\textbf{Shuchang Pan}\textsuperscript{2,*},
\textbf{Jiachen Lian}\textsuperscript{1,*,\textdagger},
\\
Siddharth Banerjee\textsuperscript{1},
Sarika Pasumarthy\textsuperscript{1}
Dhruv Hebbar\textsuperscript{1},
Siddhant Patel\textsuperscript{1},
Zeyi Austin Li\textsuperscript{1},
\\
Kan Jen Cheng\textsuperscript{1},
Sanay Bordia\textsuperscript{1}
Krish Patel\textsuperscript{1},
Akshaj Gupta\textsuperscript{1},
Tingle Li\textsuperscript{1},
Gopala Anumanchipalli\textsuperscript{1}
\\[0.5em]
\textsuperscript{1}University of California, Berkeley
\quad
\textsuperscript{2}Zhejiang University
\\
\textsuperscript{3}South China University of Technology
}

\begin{document}
\maketitle
\begin{abstract}
Human conversation is organized by an implicit chain of thought and manifests as temporally structured conversational behaviors. Capturing this perceptual pathway is critical for building natural full-duplex interactive systems. We propose S-MARC (Streaming Causal Modeling and Reasoning for Conversation), a streaming, causal, and hierarchical framework for conversational behavior modeling and reasoning. By formalizing the intent-to-action pathway, S-MARC predicts high-level communicative functions and low-level interaction behaviors while modeling their causal and temporal dependencies. To support this setting, we construct a high-quality corpus that pairs controllable, event-rich duplex dialogue data with behavior labels. S-MARC organizes streaming predictions into a continuously evolving graph structure, generates concise justifications for its decisions, and dynamically optimizes its reasoning process. Experiments on synthetic and real duplex dialogues show that S-MARC achieves robust behavior detection, produces interpretable reasoning chains, and establishes a benchmark foundation for conversational reasoning in full-duplex spoken dialogue systems. \textit{Our codes have been released in the supplementary materials and will be released on GitHub. The dataset will also be released on Hugging Face.}
\end{abstract}

\section{Introduction}
Recent advances in spoken dialogue systems have shifted from turn-based, half-duplex models to full-duplex systems capable of simultaneous listening and speaking \citep{arxiv:2503.01174, dgslm,vap}. Under this setting, the dominant paradigms frame this task as prediction. The first approach, Next Segment Prediction, models the agent's response as a complete turn \citep{hara18_interspeech, li-etal-2022-speak, 5494991}. A more recent approach, Next Dual-Token Prediction, generates simultaneous token streams for both speakers to better handle overlap and real-time interaction \citep{arxiv:2203.16502, defossez2024moshi}. While these methods have improved system responsiveness, they treat conversation as a sequence generation problem, bypassing the cognitive layer of reasoning that governs human interaction \citep{monroe2015learningrationalspeechacts,zhixuan2024pragmaticinstructionfollowinggoal}.

Human conversation, however, operates on a more abstract and causal level. When Speaker 1 produces an utterance, Speaker 2 does not simply predict the next sequence of words. Instead, they first perceive the behavior (e.g., recognizing a constative communicative function), which triggers an internal chain of thought (e.g., deciding not to interrupt and to remain silent). This reasoning process culminates in a generated action (e.g., an acknowledgement). This gap between pattern matching and reasoning is a fundamental barrier to creating truly natural AI agents. Meanwhile, full-duplex interaction imposes strict real-time requirements. When an agent is listening and speaking at the same time, it must make decisions at a per-second granularity; consequently, latency must be upper-bounded and throughput must remain stable. Our work addresses the core scientific question: under the full-duplex task setting, how can a machine model this perception-reasoning-generation loop to make principled, interpretable decisions in real time?

\begin{figure*}[t]
    \centering
\includegraphics[width=0.9\linewidth]{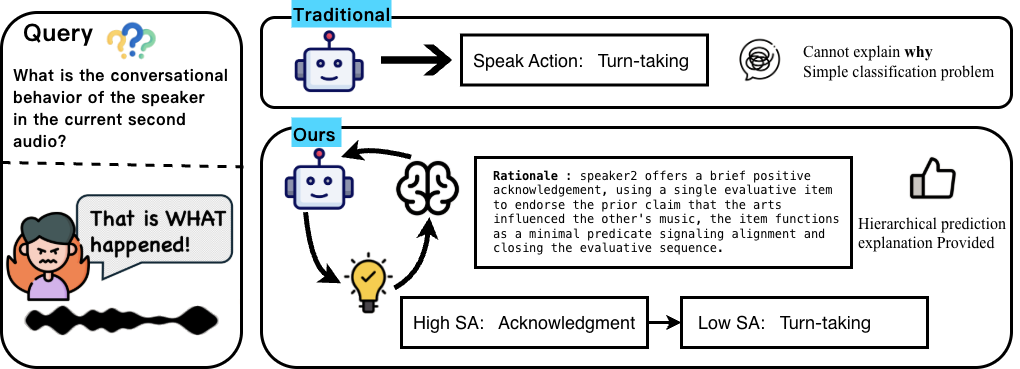}
\vspace{-5pt}
    \caption{{\bf Comparison of dialogue paradigms.} (Traditional) Traditional duplex systems frame conversation as a direct sequence prediction task. (Ours) We propose a framework based on next-behavior perception and reasoning: the agent first perceives the speaker’s behaviors at multiple levels, then reasons via a Graph-of-Thoughts, and finally generates a response.}
    \label{demo-fig-main}
\end{figure*}

To tackle this challenge, we introduce S-MARC (Streaming Causal Modeling
and Reasoning for Conversation) that operationalizes the process. Our approach is twofold. First, we formalize the Perception stage with a hierarchical conversational behavior detection model. This module learns to identify conversational behaviors at two dimensions. It first captures high-level communicative functions (e.g., {\em constative, directive}) \citep{jurafsky2025speech} that reflect coarse-grained communicative intent. Then, guided by these high-level communicative functions, it predicts low-level interaction behaviors (e.g., {\em turn-taking, backchannel}) that describe interaction mechanics \citep{schegloff1982discourse, gravano2011turn, duncan1972some, raux2012optimizing, khouzaimi2016reinforcement, marge2022spoken, arxiv:2503.04721, arxiv:2503.01174, dgslm}. This provides the system with a structured understanding of the ongoing dialogue. Second, we model the explicit reasoning process with a Graph-of-Thoughts (GoT)
system \citep{got}. Under strict real-time constraints, S-MARC treats
graph-structured memory as a low-latency surrogate for deliberative reasoning.
The system compresses past dialogue into causal evidence anchors, links these
anchors with the current tick-level behavior state, and traces a
past-to-present evidence chain to approximate the teacher model's reasoning
process. By performing inference over this graph, our model can predict the
most appropriate subsequent behavior and generate a natural-language rationale
explaining its decision. This transforms the opaque prediction task into an
auditable reasoning process, which provides a unified benchmark for
\textit{evaluating conversational behavior in duplex speech systems.}

To train this framework, we construct a high-quality hybrid corpus that combines behavior-rich dialogues with high-quality synthetic annotations that are manually verified. Experimental results show that our data effectively reproduces key interactional structures of human conversation, such as turn-taking dynamics, while the introduced annotations provide a high-fidelity characterization of conversational behaviors and the rationales underlying them.

In summary, our contributions are:
{%
\setlength{\topsep}{0pt}
\setlength{\itemsep}{0pt}
\setlength{\parskip}{0pt}
\setlength{\parsep}{0pt}
\setlength{\partopsep}{0pt} 
\begin{itemize}
    \item A conceptual shift from next-token prediction to next-behavior reasoning for full-duplex interaction, demonstrating that modeling the causal chain from intent to action is essential for natural dialogue.
    
    \item S-MARC, a streaming, causal, and hierarchical framework that jointly models high-level communicative functions, low-level interaction behaviors, and their evolving dependencies through a Graph-of-Thoughts, enabling interpretable real-time decision-making and rationale generation.
    
    \item ConversationGoT-120h, a 120-hour benchmark dataset that fills the gap left by existing dialogue resources by providing causal, one-second annotations of high-level communicative functions, low-level interaction behaviors, and evidence-based rationales over both real and synthetic duplex conversations.
\end{itemize}

\section{Related Work}

\paragraph{Spoken Dialogue Modeling.}
Recent work on incremental spoken dialogue modeling has improved real-time speech understanding and response timing, but it remains unclear how communicative functions emerge and evolve over a continuous interaction timeline, especially in full-duplex settings with overlapping speech, interruptions, floor competition, and incremental response timing. Early studies framed online dialogue understanding as low-latency semantic intent recognition before utterance completion \citep{potdar2021streamingendtoendframeworkspoken, cao2021sequential, fu2022multitask, arora2022twopass}. Because these methods mainly target task-semantic intents, they provide limited support for interactional decisions such as whether to backchannel, hold the floor, yield the turn, or interrupt. Subsequent work on multimodal online classification and utterance-level dialogue-act modeling has moved closer to communicative-function recognition \citep{miah-etal-2023-hierarchical, raheja2019dialogue, he2021speaker}. However, these methods typically rely on manually annotated utterance boundaries or predefined utterance-level units, making them difficult to apply when boundaries are uncertain and cues emerge before utterance completion. More recent studies model interaction on a continuous timeline, particularly for turn-taking and response timing \citep{arora2025talkingturnsbenchmarkingaudio, chang2022turntaking, ekstedt2022voice, kurata2023multimodal}. Yet their targets remain largely limited to low-level turn-transition phenomena rather than the communicative functions and interaction behaviors that jointly define conversational behavior states. Our work addresses this gap by modeling both high-level communicative functions and low-level interaction behaviors directly on a full-duplex interaction timeline.

\paragraph{Faithful Rationales.}
In online dialogue modeling, intermediate conversational behavior labels can support downstream decisions, but discrete labels alone do not reveal the contextual evidence behind each prediction. Prior work on spoken dialogue state tracking and incremental dialogue management highlights the value of explicit, updateable intermediate states for policy selection and uncertainty handling \citep{williams2012challenges,metallinou-etal-2013-discriminative,44018,inproceedings,lee2021dialoguestatetrackinglanguage,kennington2025priorlessonsincrementaldialogue}. Recent studies have further explored natural-language explanations and speech-based emotion reasoning by combining large language models with acoustic cues \citep{wang2026emotionthinkerprosodyawarereinforcementlearning,zhang2025classificationspeechemotionreasoning,xu2023secapspeechemotioncaptioning}. However, generating open-ended rationales with large models at every decision step incurs substantial latency, limiting their use in real-time systems. We therefore seek faithful, lightweight rationales that expose the evidence behind behavior-state predictions while preserving streaming efficiency.

\section{ConversationGoT-120h}
ConversationGoT-120h is a 120-hour causal streaming dataset for second-level conversational behavior modeling. It contains real and synthetic two-person open-domain conversations, with each one-second segment labeled by a hierarchical behavior state consisting of a communicative function, an interaction behavior, and an evidence-based rationale. All annotations are produced using only past and current context.

\begin{table}[t]
\centering

{\fontsize{7}{8}\selectfont
\setlength{\tabcolsep}{3pt}
\renewcommand{\arraystretch}{1.0}

\begin{adjustbox}{width=\columnwidth}
\begin{tabular}{l C{0.8cm} C{1.25cm} C{0.7cm} C{1.15cm} C{1.15cm} C{0.9cm}}
\toprule
\textbf{Benchmark}
& \textbf{Hours}
& \makecell[c]{\textbf{Label}\\\textbf{gran.}}
& \textbf{Causal}
& \makecell[c]{\textbf{Comm.}\\\textbf{Func.}}
& \makecell[c]{\textbf{Inter.}\\\textbf{Behav.}}
& \textbf{Rat.} \\
\midrule
HCRC Map Task~\cite{Thompson1993MapTask}
& 18 h
& Utt.
& \xmark
& \xmark
& \cmark
& \xmark \\

Candor~\cite{Reece2023Candor}
& 850+ h
& Conv.
& \xmark
& \xmark
& \cmark
& \xmark \\

SwDA~\cite{Stolcke2000DialogueAct}
& 96.3 h
& Utt.
& \xmark
& \cmark
& \xmark
& \xmark \\

RASwDA~\cite{Chen2024RASwDA}
& 44.8 h
& Utt.
& \xmark
& \cmark
& \xmark
& \xmark \\

MRDA~\cite{Shriberg2004MRDA}
& 72 h
& Utt.
& \xmark
& \cmark
& \xmark
& \xmark \\

\makecell[l]{\textbf{Conversation}\\\textbf{GoT-120h}}
& 120 h
& \textbf{1 s}
& \cmark
& \cmark
& \cmark
& \cmark \\
\bottomrule
\end{tabular}
\end{adjustbox}
}

\caption{Comparison of relevant public resources by output granularity and label structure.}
\label{tab:benchmark_comparison}
\end{table}
\paragraph{Conversational Behavior State and Rationale Taxonomy.}
As shown in Table~\ref{tab:benchmark_comparison}, existing dialogue resources typically provide either utterance-level communicative-function labels or coarse interaction-behavior annotations, but rarely offer fine-grained causal labels that jointly capture communicative intent and local interaction dynamics. To address this gap, we define a two-dimensional conversational behavior state taxonomy: high-level communicative functions and low-level interaction behaviors. The complete Conversational Behavior State taxonomy, including label definitions and decision rules, is provided in the Appendix.

\textbf{(i) Low-level interaction behaviors} capture real-time turn-management roles, such as turn holding, turn taking, interruption, and backchannel feedback, following conversation-analysis and psycholinguistic studies of turn-taking~\cite{Sacks1974TurnTaking,Stivers2009TurnTaking,Levinson2015Timing}.
\textbf{(ii) High-level communicative functions} capture pragmatic intent, following dialogue-act annotation schemes that label utterances by actions such as informing, questioning, requesting, committing, and acknowledging~\cite{Stolcke2000DialogueAct,ISO24617-2,Bunt2020ISO}.
\textbf{(iii) Rationale (Rat.)} makes each annotation evidence-grounded and auditable, following prior work on annotator rationales and rationalized NLP~\cite{Zaidan2007Rationales,DeYoung2020ERASER}. It explains how the segment functions in the local interaction and what contextual cues support the assigned communicative function and interaction behavior.

\paragraph{Data Composition and Annotation Pipeline.}

ConversationGoT-120h consists of 60 hours of real conversations and 60 hours of synthetic conversations. The real subset is selected from Candor for high-quality two-person open-domain interactions, while the synthetic subset is designed to improve coverage of speaker identities, topics, acoustic conditions, and conversational styles. Annotations are produced in a causal streaming manner: at each timestamp, the annotator observes only the current segment and past context, retrieves relevant historical anchors, and predicts the behavior state with an evidence-based rationale. The retrieved anchors further provide supervision for learning evidence-grounded causal reasoning chains in streaming dialogue. We report annotation quality checks in the experimental section, and provide pipeline details in the Appendix.

\section{S-MARC}

\begin{figure*}[h]
    \centering
    \includegraphics[width=0.9\linewidth]{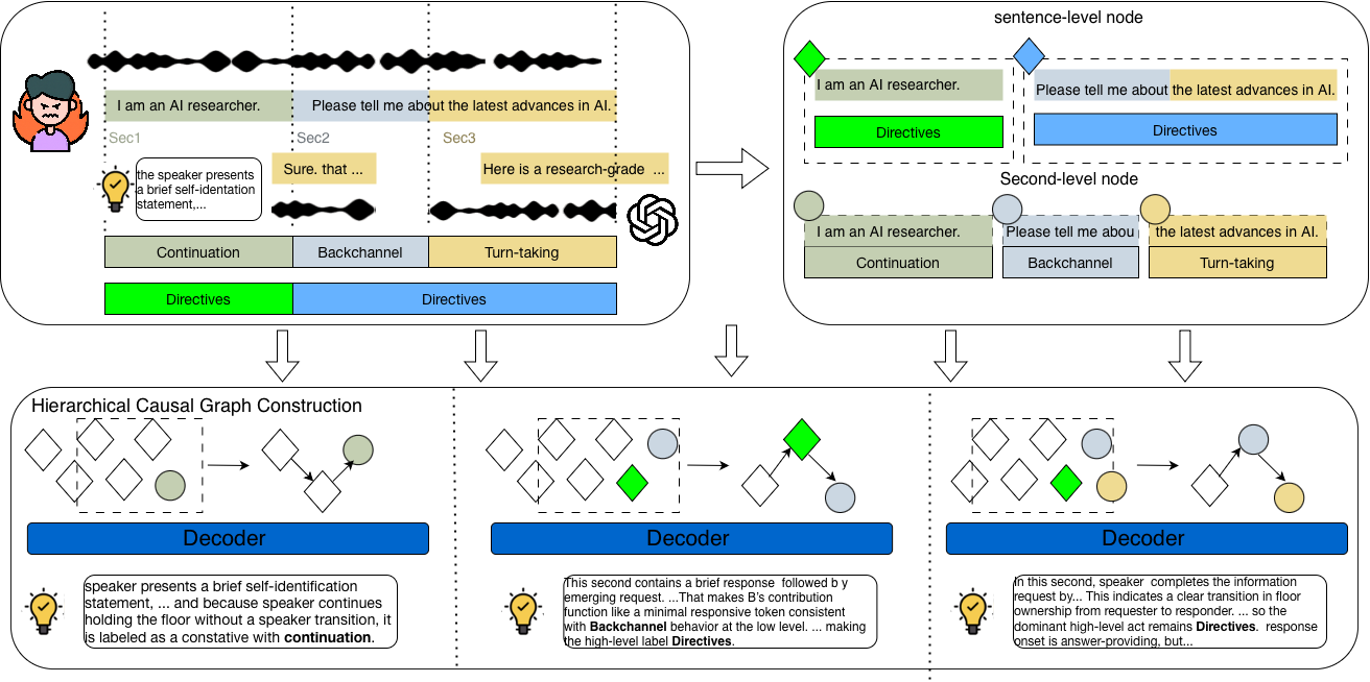}
    \vspace{-5pt}
    \caption{{\bf Causal streaming pipeline for conversational behavior modeling.} At each 1~s tick, S-MARC predicts conversational behavior states and generates evidence-grounded rationales with a sliding-window Graph-of-Thoughts (GoT).}
    \label{pipeline}
\end{figure*}

To address the full-duplex conversational behavior modeling challenges introduced earlier, we propose S-MARC, a causal streaming framework for low-latency behavior perception and rationale generation. As shown in Figure~\ref{pipeline}, S-MARC maintains an online-updated causal node chain. A Conversational Behavior State perceiver predicts behavior states at each tick, while a GoT decoder retrieves historical evidence and generates grounded rationales. We first define the causal node chain and its online update procedure, then describe the training process.

\subsection{Causal Node Chain and Online Update}
S-MARC maintains a memory graph
$\mathcal{M}_t=(\mathcal{V}_t,\mathcal{E}_t)$ at each one-second timestamp
$t$. The graph stores retained information available up to the current timestamp
instead of the full dialogue history. It contains tick-level nodes for fine-grained
streaming observations and sentence-level nodes for compact long-range discourse
memory.

\paragraph{Tick-Level Node.}

A tick-level node records the observation at the current second:
$
v_t^{\mathrm{tick}}=\langle a_t,x_t,\rho_t,b_t\rangle ,
$
where $a_t$ is the current audio segment, $x_t$ is the incremental transcript,
$\rho_t$ is the timestamp metadata, and $b_t$ is the predicted conversational
behavior state, including both the high-level communicative function and the
low-level interaction behavior.

\paragraph{Sentence-Level Node.}

When a sentence boundary is confirmed, the covered tick-level nodes are
compressed into a sentence-level node:
\[
v_k^{\mathrm{sent}}
=
\langle X_k,[\tau_k^{s},\tau_k^{e}),R_k,B_k\rangle .
\]
Here, $X_k$ is the committed sentence text, $[\tau_k^{s},\tau_k^{e})$ is its
time span, $R_k$ is the merged timestamp metadata, and $B_k$ collects the
predicted behavior states of the covered ticks. Sentence-level nodes serve as
compact anchors for later reasoning.

\paragraph{Causal Edges.}

The edge set connects retained nodes chronologically. Each edge points from an
older retained node to the next newer retained node, forming a past-to-present
causal chain. When sentence-level nodes are inserted or removed, the chain is
updated accordingly without introducing additional edge types.

\paragraph{Online Update.}

At each timestamp $t$, S-MARC updates the memory graph in four steps. First,
expired sentence-level nodes are pruned. Given a memory window $W$, any
sentence node whose end time is earlier than $t-W$ is removed with its incident
edges. Second, the new audio segment is processed by the causal ASR module:
$
x_t,\rho_t=\operatorname{ASR}(a_t).
$
S-MARC creates a new tick-level node
$
v_t^{\mathrm{tick}}=\langle a_t,x_t,\rho_t,\varnothing\rangle
$
and inserts it into the active local buffer $\mathcal{Q}_t$, with the behavior
state initialized as empty before prediction.

Third, sentence-boundary checking is applied only to the stable prefix of the
buffer, excluding the latest tick-level node. If a boundary is detected, the
completed span is committed as a sentence-level node. Only tick-level nodes
before the current timestamp with predicted behavior states are folded into the
sentence node, while the latest tick-level node remains active. Finally, the new
sentence-level node, if any, is inserted into $\mathcal{M}_t$, and the
chronological causal chain is rebuilt over all retained nodes. This allows
downstream reasoning to use compact historical sentence memory and fine-grained
current tick observations while preserving causal online updates.

\subsection{Conversational Behavior State Perception}
\begin{table}[t]
\centering
\caption{Empirical distribution of low-level interaction behaviors conditioned on high-level communicative functions.}
\label{tab:cond_behavior_distribution}

{\fontsize{7.5}{9.5}\selectfont
\renewcommand{\arraystretch}{0.92} 
\setlength{\tabcolsep}{3pt}     

\begin{tabular}{@{}l|cccc@{}}
\toprule
Function & Continuation & Turn-taking & Interruption & Backchannel \\
\midrule
Constatives      & \textbf{0.5375} & 0.2280 & 0.1320 & 0.1025 \\
Directives       & \textbf{0.4903} & 0.2409 & 0.1619 & 0.1069 \\
Acknowledgments  & 0.2552 & 0.2325 & 0.1593 & \textbf{0.3530} \\
Commissives      & \textbf{0.4723} & 0.2620 & 0.1393 & 0.1264 \\
\bottomrule
\end{tabular}
}
\end{table}

After obtaining the updated causal chain $\mathcal{M}_t=(\mathcal{V}_t,\mathcal{E}_t)$, S-MARC injects conversational-behavior-state awareness into the latest tick-level node $v_t^{\mathrm{tick}}$. Unlike standard frame-wise classification, this task requires strictly causal tick-level inference, with uncertainty mainly around state boundaries. As shown in Table~\ref{tab:cond_behavior_distribution}, high-level communicative functions and low-level interaction behaviors are statistically coupled but evolve at different temporal granularities. S-MARC therefore adopts a hierarchical streaming decoder, where high-level communicative functions guide low-level interaction-behavior prediction, and the low-level interaction behavior is modeled as a duration-aware evolving belief.

\paragraph{Causal Evidence Construction.}
Given the dual-stream features $(h_t^B, h_t^E)$ from frozen encoders, where the ASR-text stream is produced by NVIDIA Nemotron Speech Streaming EN 0.6B and the acoustic stream is extracted from the model's frozen encoder, S-MARC fuses the acoustic and ASR-text streams with a learnable gate and projects them into a shared tick representation $z_t$. The causal evidence trunk then extracts transition-sensitive evidence using only prefix information:
\[
e_t
=
\phi_{\mathrm{ev}}
\left(
z_t,\,
z_{t-1},\,
z_t-z_{t-1},\,
\mathrm{CausalConv}(z_{1:t})
\right),
\]
where $z_{t-1}=z_0$ for $t=0$. The terms $z_t$ and $z_{t-1}$ encode current and
previous tick representations, $z_t-z_{t-1}$ highlights local changes, and
$\mathrm{CausalConv}(z_{1:t})$ summarizes prefix-only temporal context.

\paragraph{High-level Communicative-Function Guidance.}
S-MARC first predicts the high-level communicative function from $e_t$ and
derives a continuous guidance vector for low-level interaction-behavior decoding:
$
\ell_t^{H},\,g_t
=
\phi_H(e_t),
$
where $\ell_t^{H}\in\mathbb{R}^{|\mathcal{Y}_H|}$ denotes high-level
communicative-function logits and $g_t\in\mathbb{R}^{d_g}$ denotes
communicative-function guidance. Rather than serving as rule-based
post-processing, $g_t$ acts as a learnable conditioning signal for low-level
interaction-behavior inference.

\paragraph{Duration-conditioned Belief-state Decoding.}
To capture the persistence of low-level interaction behavior and reduce
tick-wise jitter, S-MARC maintains a duration-conditioned belief distribution
$\pi_t\in\Delta^{|\mathcal{Y}_L|\times|\mathcal{D}|}$ over the low-level
interaction-behavior label space $\mathcal{Y}_L$ and the duration-bucket space
$\mathcal{D}$.

At each tick, the decoder uses the current evidence $e_t$, high-level guidance
$g_t$, and previous belief $\pi_{t-1}$ to estimate a switch gate
$\alpha_t\in[0,1]$ and a candidate next-label distribution
$q_t\in\Delta^{|\mathcal{Y}_L|}$. The belief is updated by a duration-aware
transition:
\[
\begin{aligned}
\pi_t
&=
(1-\alpha_t)\,\mathrm{Shift}(\pi_{t-1})
+
\alpha_t\,\mathrm{Reset}(q_t),\\
\ell_t^{L}
&=
\phi_L(e_t,g_t,\pi_t).
\end{aligned}
\]
Here, $\mathrm{Shift}(\cdot)$ advances the duration bucket with saturation when
the current state is preserved, while $\mathrm{Reset}(\cdot)$ places the
candidate distribution $q_t$ in the initial duration bucket when a state switch
is detected. The updated belief $\pi_t$ is combined with $e_t$ and $g_t$ to
produce the low-level interaction-behavior logits
$\ell_t^{L}\in\mathbb{R}^{|\mathcal{Y}_L|}$. This update preserves the current interaction behavior when switch evidence is
weak and resets the belief when boundary evidence is strong, reducing
fragmentation while retaining sensitivity to short-lived behaviors such as
Backchannel and Interruption. After decoding, S-MARC derives the current
conversational behavior state $b_t$ from the high-level and low-level
predictions and writes it back to the latest tick-level node:
$
v_t^{\mathrm{tick}}
\leftarrow
\langle a_t,x_t,\rho_t,b_t\rangle .
$

\subsection{GoT Rationale Generation}
Given the updated memory graph and the current behavior-aware tick-level node,
S-MARC distills the teacher model's evidence-grounded reasoning into a
low-latency rationale. The GoT decoder retrieves compact historical evidence,
integrates local tick-level context, and organizes the selected nodes into a
past-to-present reasoning chain under causal and real-time constraints.

\paragraph{Evidence-chain Selection and Local Context Augmentation}

At timestamp $t$, the GoT decoder receives the updated causal memory graph
$\mathcal{M}_t$, the current tick-level node $v_t^{\mathrm{tick}}$, and the
retained sliding-window memory. We use $v_t^{\mathrm{tick}}$ as the query node
$q_t$ and collect committed sentence-level nodes within the causal window as
the candidate evidence set $\mathcal{C}_t$. For each candidate sentence node
$v_k^{\mathrm{sent}}\in\mathcal{C}_t$, the selector computes a relevance score
$s_{t,k}$ conditioned on $q_t$ and $\mathcal{M}_t$. In parallel, a threshold
predictor estimates a query-dependent threshold $\tau_t$, allowing the number
of selected evidence nodes to adapt to the dialogue context. The selected evidence chain is obtained by retaining candidate sentence nodes
whose scores exceed $\tau_t$ and sorting them chronologically. We denote the
resulting ordered evidence chain as $\hat{\mathcal{A}}_t$. Because
sentence-level anchors may miss local acoustic or turn-management cues, S-MARC
further retrieves recent committed sentence nodes $R_t$ and active
within-sentence tick-level nodes $P_t$. The final evidence context is
$
\xi_t=(\hat{\mathcal{A}}_t,R_t,P_t,q_t),
$
where $\hat{\mathcal{A}}_t$ captures global historical evidence and
$(R_t,P_t,q_t)$ provides local observable details.

\paragraph{Past-to-present Chain Decoding}

Given $\xi_t$, S-MARC sorts the evidence nodes and the current query node by
timestamp and connects adjacent nodes chronologically, forming a causal
past-to-present reasoning chain as the Graph-of-Thoughts context. Instead of applying an additional graph network, S-MARC linearizes the chain
into a token sequence and feeds it to a sequence-to-sequence decoder. The
rationale is generated autoregressively:
\[
p_\theta(r_t\mid \xi_t)
=
\prod_{n=1}^{|r_t|}
p_\theta(r_{t,n}\mid r_{t,<n},\xi_t),
\]
where $r_t$ denotes the rationale for the current tick. Since $\xi_t$ is constructed only from the retained causal window, the generated
rationale cannot access future observations. Thus, each explanation is grounded
in selected historical evidence and local tick-level context, making the GoT
decoder consistent with streaming deployment.

\subsection{Training Objective}

S-MARC is trained with behavior-state supervision and teacher-supervised GoT
rationale supervision. The behavior perceiver uses class-balanced cross-entropy
for high-level communicative-function prediction and class-balanced focal loss
for low-level interaction-behavior prediction:
\[
\mathcal{L}_{\mathrm{beh}}
=
\lambda_H \mathcal{L}_{H}^{\mathrm{CE}}
+
\lambda_L \mathcal{L}_{L}^{\mathrm{focal}} .
\]
The GoT module is optimized with an evidence-selection loss and a
teacher-forced rationale generation loss:
\[
\mathcal{L}_{\mathrm{GoT}}
=
\lambda_{\mathrm{sel}}\mathcal{L}_{\mathrm{sel}}
+
\lambda_{\mathrm{dec}}\mathcal{L}_{\mathrm{dec}} .
\]
Detailed loss definitions and training configurations are provided in
Appendix.

\section{Experiments}
\subsection{ConversationGoT-120h}

We split ConversationGoT-120h into training, validation, and test sets with a 6:2:2 ratio.

\paragraph{Quality Check.}
Dialogue quality directly affects the reliability of downstream behavior-state labels and rationale annotations. We therefore conduct a two-stage quality check. First, we manually filter low-quality or low-authenticity synthetic dialogues. Second, five human volunteers and GPT-4o independently rate the remaining samples along predefined quality dimensions. Table~\ref{tab:human_vs_gpt4o_subjective} reports the subjective quality of dialogue content. Human volunteers give conservative but generally positive ratings, with Consistency reaching $7.98\pm1.12$ and the other dimensions ranging from $6.36$ to $6.71$. GPT-4o assigns higher scores across most dimensions, suggesting that automatic evaluation is more optimistic than human judgment. We therefore treat human ratings as the primary quality signal and use GPT-4o ratings as a complementary reference. The rationale annotations are used as supervision for GoT, including evidence-chain construction and rationale generation conditioned on selected evidence. We further evaluate rationale quality along six dimensions, as shown in Table~\ref{tab:rationale_quality_subjective}. Human ratings range from $6.79$ to $8.09$, while GPT-4o ratings range from $7.62$ to $8.28$. The two evaluators show broadly consistent trends, although GPT-4o again tends to assign higher scores and larger variances in some dimensions. These results suggest that the rationales are generally plausible, context-grounded, and coherent, while leaving room for more conservative human-side validation.

\begin{table}[t]
\centering
\caption{Subjective comparison between human volunteers and GPT-4o across four conversational quality dimensions.}
\label{tab:human_vs_gpt4o_subjective}
{\fontsize{7}{8}\selectfont
\renewcommand{\arraystretch}{1.0}
\setlength{\tabcolsep}{3pt}
\begin{tabular}{@{}lcccc@{}}
\toprule
\textbf{Rater} 
& \textbf{Naturalness} 
& \textbf{Consistency} 
& \makecell[c]{\textbf{Reasonableness}} 
& \textbf{Human-Likeness} \\
\midrule
Volunteers 
& $6.71 \pm 1.32$ 
& $7.98 \pm 1.12$ 
& $6.36 \pm 0.90$ 
& $6.46 \pm 0.92$ \\
GPT-4o     
& $8.34 \pm 1.35$ 
& $8.10 \pm 1.34$ 
& $8.42 \pm 0.93$ 
& $7.64 \pm 0.90$ \\
\bottomrule
\end{tabular}
}
\end{table}

\begin{table}[t]
\centering
\caption{Subjective comparison between human volunteers and GPT-4o across six rationale quality dimensions.}
\label{tab:rationale_quality_subjective}
{\fontsize{6.5}{7.2}\selectfont
\renewcommand{\arraystretch}{1.0}
\setlength{\tabcolsep}{3pt}
\begin{tabular}{@{}lccc@{}}
\toprule
\textbf{Rater} 
& \textbf{Reasonableness} 
& \makecell[c]{\textbf{Context}\\\textbf{Grounding}} 
& \makecell[c]{\textbf{Intra-Utterance}\\\textbf{Coherence}} \\
\midrule
Volunteers 
& $7.85 \pm 1.02$ 
& $7.88 \pm 0.86$ 
& $6.79 \pm 1.13$ \\
GPT-4o     
& $7.62 \pm 2.31$ 
& $8.28 \pm 1.49$ 
& $7.88 \pm 1.88$ \\
\midrule
\textbf{Rater} 
& \makecell[c]{\textbf{Inter-Utterance}\\\textbf{Coherence}} 
& \makecell[c]{\textbf{Specificity}\\\textbf{\& Focus}} 
& \makecell[c]{\textbf{Clarity \&}\\\textbf{Non-Template Style}} \\
\midrule
Volunteers 
& $7.61 \pm 0.94$ 
& $7.72 \pm 0.87$ 
& $8.09 \pm 1.00$ \\
GPT-4o     
& $8.13 \pm 1.69$ 
& $7.89 \pm 1.84$ 
& $8.00 \pm 1.69$ \\
\bottomrule
\end{tabular}
}
\end{table}

\paragraph{Dataset Statistics.}
We provide additional dataset diagnostics in the Appendix, including overlap and interaction statistics, event-label distributions, and anchor statistics. These analyses show that ConversationGoT-120h contains dense turn-level interaction signals while exhibiting natural label imbalance across both high-level communicative functions and low-level interaction behaviors.

\subsection{Conversational Behavior State Perceiver}
We adopt a three-pronged evaluation protocol: 
(i) comparisons with prior dialogue-act work and low-level interaction-behavior prediction work; 
(ii) ablation studies of our proposed model; and 
(iii) human judgments of per-second decisions to assess perceptual validity and capture aspects missed by automated metrics. 
Detailed metric definitions are provided in the Appendix.

\begin{table*}[h]
\centering
\caption{per-class F1/AUC scores and inference latency for high-level communicative functions and low-level interaction behaviors. Each class cell reports F1 / AUC. Inference latency is reported as the mean per-sample inference time in seconds.}
\label{tab:simulation_results_detection}
\scriptsize
\resizebox{\textwidth}{!}{
\begin{tabular}{lcccccccccc}
\toprule
\textbf{Method}
& \textbf{H-Const.} 
& \textbf{H-Dir.} 
& \textbf{H-Comm.} 
& \textbf{H-Ack.}
& \textbf{L-Sil.}
& \textbf{L-Turn.} 
& \textbf{L-Inter.} 
& \textbf{L-Back.} 
& \textbf{L-Cont.}
& \textbf{Infer. (s)} \\
\midrule
TalkingTurns\citep{arora2025talkingturns}
& NA
& NA
& NA
& NA
& 0.083 / 0.713
& 0.063 / 0.601
& 0.196 / 0.639
& 0.029 / 0.540
& 0.699 / 0.892
& 2.185 \\

MIDAS \citep{yu2021midas}
& 0.799 / 0.723
& 0.397 / 0.823
& NA
& 0.172 / 0.692
& NA
& NA
& NA
& NA
& NA
& NA \\
\midrule
NoTrunk
& 0.732 / 0.552
& 0.098 / 0.522
& 0.133 / 0.627
& 0.090 / 0.499
& 0.294 / 0.854
& 0.145 / 0.495
& 0.028 / 0.533
& 0.033 / 0.415
& 0.369 / 0.483
& 0.100 \\

FixedAverageFusion
& 0.788 / 0.768
& 0.432 / 0.794
& 0.369 / 0.864
& 0.433 / 0.759
& 0.656 / 0.959
& 0.475 / 0.818
& 0.102 / 0.760
& 0.314 / 0.732
& 0.776 / 0.812
& 0.099 \\

NoBeliefFilter
& 0.785 / 0.790
& 0.430 / 0.785
& 0.380 / 0.840
& 0.435 / 0.765
& 0.625 / 0.940
& 0.405 / 0.765
& 0.085 / 0.710
& 0.265 / 0.690
& 0.720 / 0.775
& 0.045 \\

NoHighGuidanceToLow
& 0.800 / 0.805
& 0.475 / 0.815
& 0.415 / 0.860
& 0.485 / 0.785
& 0.665 / 0.955
& 0.485 / 0.825
& 0.170 / 0.775
& 0.380 / 0.760
& 0.755 / 0.815
& 0.100 \\

\midrule
Ours 
& 0.828 / 0.841
& 0.534 / 0.861
& 0.462 / 0.893
& 0.546 / 0.837
& 0.707 / 0.975
& 0.578 / 0.883
& 0.386 / 0.851
& 0.495 / 0.833
& 0.824 / 0.885
& 0.101 \\
\bottomrule
\end{tabular}
}
\end{table*}

\paragraph{Baseline Comparison}
For the high-level baseline, we evaluate MIDAS by mapping its original dialogue-act labels to our high-level communicative-function categories, with the full mapping reported in the Appendix; H-Comm. is left unmapped because MIDAS has no corresponding category. For the low-level baseline, we evaluate TalkingTurns predictions after aligning its output labels with our low-level interaction-behavior definitions. As shown in Table~\ref{tab:simulation_results_detection}, our method outperforms the external high-level baseline MIDAS on all comparable communicative functions, with the largest gain on H-Ack. from 0.172/0.692 to 0.546/0.837. This suggests that the hierarchical perceiver better captures categories requiring local interaction context and causal-temporal cues. Compared with the low-level baseline TalkingTurns, our method improves most fine-grained behaviors, notably L-Sil., L-Turn., and L-Back., while remaining competitive on L-Cont. at 0.824/0.885. It also reduces inference time from TalkingTurns’s 2.185s to 0.101s and jointly covers both high- and low-level tasks. These results indicate that the proposed framework provides more accurate and efficient joint modeling of communicative functions and interaction behaviors.

\paragraph{Ablation Studies.}
We evaluate each component under the same input and training protocol by comparing the full model with four ablations: NoBeliefFilter removes the low-level recurrent belief filter, NoTrunk removes the CausalEvidenceTrunk, NoHighGuidanceToLow disables high-level guidance to the low-level heads while retaining high-level supervision, and FixedAverageFusion replaces learned fusion with fixed averaging. As shown in Table~\ref{tab:simulation_results_detection}, the full model consistently outperforms all ablated variants on both high-level and low-level behavior prediction, with inference latency comparable to most alternatives. The largest degradation occurs when removing the CausalEvidenceTrunk: H-Dir. F1 drops from 0.534 to 0.098 and L-Cont. F1 from 0.824 to 0.369, despite nearly identical latency, highlighting its role in extracting causal evidence for temporal decision-making. Replacing learned fusion with fixed averaging also reduces performance, especially on H-Dir. and L-Back., with almost no latency benefit, suggesting that adaptive fusion better coordinates belief and evidence representations. Removing the belief filter substantially lowers low-level performance, particularly on L-Inter. and L-Back., indicating that recurrent belief states are important for short-term, state-dependent interactive behaviors. Finally, removing high-to-low guidance degrades both H-Ack. and L-Turn. with little latency change, showing that high-level guidance improves low-level transition modeling at negligible cost. Overall, these results demonstrate that the full architecture provides a better trade-off among accuracy, hierarchical consistency, and inference efficiency.

\paragraph{Human--Model Agreement.}
To validate whether the predicted conversational behavior states are behaviorally plausible, we report Human--Model Agreement (HMA) with both human volunteers and GPT-4o as raters. The annotation protocol and formal definition of HMA are provided in the Appendix. As shown in Table~\ref{tab:hma_human_vs_gpt4o}, the predicted labels receive high agreement from both groups. Human volunteers show particularly strong agreement on high-level communicative functions (HMA$^{h}=0.97$), while GPT-4o shows stronger agreement on low-level interaction behaviors (HMA$^{l}=0.93$). The lower human agreement on low-level labels (HMA$^{l}=0.77$) likely reflects boundary ambiguity in second-level behaviors, such as turn-taking, backchannels, and interruptions. We therefore treat human ratings as the primary validation signal and GPT-4o ratings as a complementary automatic reference.
\begin{table}[t]
\centering
\caption{Human--Model Agreement (HMA) for high-level and low-level behavior-state predictions.}
\label{tab:hma_human_vs_gpt4o}
{\fontsize{7}{7.5}\selectfont
\setlength{\tabcolsep}{6pt}
\renewcommand{\arraystretch}{1.05}
\resizebox{0.7\linewidth}{!}{
\begin{tabular}{lcc}
\toprule
\textbf{Rater} & \textbf{HMA$^{h}$ ($\uparrow$)} & \textbf{HMA$^{l}$ ($\uparrow$)} \\
\midrule
Volunteers & 0.97 & 0.77 \\
GPT-4o & 0.78 & 0.93 \\
\bottomrule
\end{tabular}
}
}
\end{table}

\subsection{Graph of Thought}

We evaluate rationale quality and generation latency under a controlled setting where all methods receive the same input context and are evaluated by the same automatic judge. We compare four methods: (i) our GoT method; (ii) a random-selector baseline, which keeps the same GoT reasoner but replaces selected anchors with $k\in[2,8]$ uniformly sampled candidates; (iii) GPT-4o as a lightweight LLM baseline; and (iv) GPT-5 (thinking) as a stronger reasoning baseline. Gemini 2.5 Pro is used as the automatic judge with a fixed Ruler rubric covering four dimensions: Alignment, Justification, Caption Completeness, and Clarity. We report both inference latency and rationale quality scores. As shown in Table~\ref{tab:ruler_latency}, our method substantially outperforms the random-selector baseline across all Ruler dimensions with nearly identical latency, showing that evidence-anchor selection is critical for generating consistent and traceable rationales. Compared with directly prompting GPT-4o, our method is faster ($0.74$\,s vs. $2.98$\,s) and achieves higher scores on all four dimensions. GPT-5 (thinking) obtains the highest overall quality scores, but requires $16.98$\,s on average, making it more than $20\times$ slower than our method. These results show that GoT offers a favorable quality--latency trade-off for full-duplex settings that require high-frequency rationale generation.

\begin{table}[t]
\centering
\caption{Reasoning latency and subjective scores across methods.}
\label{tab:ruler_latency}
{\fontsize{9}{7.5}\selectfont
\setlength{\tabcolsep}{6pt}
\renewcommand{\arraystretch}{1.10}
\resizebox{\linewidth}{!}{
\begin{tabular}{l c cccc}
\toprule
\textbf{Method} & \textbf{Latency (s)} 
& \textbf{Alignment} & \textbf{Justification} & \textbf{Caption} & \textbf{Clarity} \\
\midrule
Ours & $0.74 \pm 0.12$ & 4.31 & 4.42 & 4.03 & 4.50 \\
Random selector & $0.73 \pm 0.11$ & 3.38 & 3.04 & 3.29 & 3.75 \\
GPT-4o & $2.98 \pm 1.04$ & 3.52 & 3.16 & 3.30 & 3.24 \\
GPT-5 (thinking) & $16.98 \pm 5.29$ & 4.60 & 4.18 & 4.43 & 4.78 \\
\bottomrule
\end{tabular}
}}
\end{table}

\section{Conclusion}
We presented a causal streaming framework for full-duplex, per-second conversational behavior state recognition. By modeling function-to-behavior reasoning with Graph-of-Thoughts, the framework also generates auditable natural-language explanations. Experimental results demonstrate stable recognition performance, improved explanation quality over random evidence chains, and low latency for real-time full-duplex interaction.

\section*{Limitations}

S-MARC must balance real-time full-duplex decision-making with the computational complexity of graph-structured reasoning, which limits the depth, size, and update frequency of the Graph-of-Thoughts used during streaming inference. Future work can explore more expressive, efficient, and generalizable graph structures that preserve real-time responsiveness while improving reasoning capacity across diverse dialogue scenarios.

\nocite{openai2024gpt4ocard,singh2025openaigpt5card,panayotov2015librispeech,du2024cosyvoice}
\bibliography{custom}

@misc{singh2025openaigpt5card,
      title={OpenAI GPT-5 System Card}, 
      author={Aaditya Singh and Adam Fry and Adam Perelman and Adam Tart and Adi Ganesh and Ahmed El-Kishky and Aidan McLaughlin and Aiden Low and AJ Ostrow and Akhila Ananthram and Akshay Nathan and Alan Luo and Alec Helyar and Aleksander Madry and Aleksandr Efremov and Aleksandra Spyra and Alex Baker-Whitcomb and Alex Beutel and Alex Karpenko and Alex Makelov and Alex Neitz and Alex Wei and Alexandra Barr and Alexandre Kirchmeyer and Alexey Ivanov and Alexi Christakis and Alistair Gillespie and Allison Tam and Ally Bennett and Alvin Wan and Alyssa Huang and Amy McDonald Sandjideh and Amy Yang and Ananya Kumar and Andre Saraiva and Andrea Vallone and Andrei Gheorghe and Andres Garcia Garcia and Andrew Braunstein and Andrew Liu and Andrew Schmidt and Andrey Mereskin and Andrey Mishchenko and Andy Applebaum and Andy Rogerson and Ann Rajan and Annie Wei and Anoop Kotha and Anubha Srivastava and Anushree Agrawal and Arun Vijayvergiya and Ashley Tyra and Ashvin Nair and Avi Nayak and Ben Eggers and Bessie Ji and Beth Hoover and Bill Chen and Blair Chen and Boaz Barak and Borys Minaiev and Botao Hao and Bowen Baker and Brad Lightcap and Brandon McKinzie and Brandon Wang and Brendan Quinn and Brian Fioca and Brian Hsu and Brian Yang and Brian Yu and Brian Zhang and Brittany Brenner and Callie Riggins Zetino and Cameron Raymond and Camillo Lugaresi and Carolina Paz and Cary Hudson and Cedric Whitney and Chak Li and Charles Chen and Charlotte Cole and Chelsea Voss and Chen Ding and Chen Shen and Chengdu Huang and Chris Colby and Chris Hallacy and Chris Koch and Chris Lu and Christina Kaplan and Christina Kim and CJ Minott-Henriques and Cliff Frey and Cody Yu and Coley Czarnecki and Colin Reid and Colin Wei and Cory Decareaux and Cristina Scheau and Cyril Zhang and Cyrus Forbes and Da Tang and Dakota Goldberg and Dan Roberts and Dana Palmie and Daniel Kappler and Daniel Levine and Daniel Wright and Dave Leo and David Lin and David Robinson and Declan Grabb and Derek Chen and Derek Lim and Derek Salama and Dibya Bhattacharjee and Dimitris Tsipras and Dinghua Li and Dingli Yu and DJ Strouse and Drew Williams and Dylan Hunn and Ed Bayes and Edwin Arbus and Ekin Akyurek and Elaine Ya Le and Elana Widmann and Eli Yani and Elizabeth Proehl and Enis Sert and Enoch Cheung and Eri Schwartz and Eric Han and Eric Jiang and Eric Mitchell and Eric Sigler and Eric Wallace and Erik Ritter and Erin Kavanaugh and Evan Mays and Evgenii Nikishin and Fangyuan Li and Felipe Petroski Such and Filipe de Avila Belbute Peres and Filippo Raso and Florent Bekerman and Foivos Tsimpourlas and Fotis Chantzis and Francis Song and Francis Zhang and Gaby Raila and Garrett McGrath and Gary Briggs and Gary Yang and Giambattista Parascandolo and Gildas Chabot and Grace Kim and Grace Zhao and Gregory Valiant and Guillaume Leclerc and Hadi Salman and Hanson Wang and Hao Sheng and Haoming Jiang and Haoyu Wang and Haozhun Jin and Harshit Sikchi and Heather Schmidt and Henry Aspegren and Honglin Chen and Huida Qiu and Hunter Lightman and Ian Covert and Ian Kivlichan and Ian Silber and Ian Sohl and Ibrahim Hammoud and Ignasi Clavera and Ikai Lan and Ilge Akkaya and Ilya Kostrikov and Irina Kofman and Isak Etinger and Ishaan Singal and Jackie Hehir and Jacob Huh and Jacqueline Pan and Jake Wilczynski and Jakub Pachocki and James Lee and James Quinn and Jamie Kiros and Janvi Kalra and Jasmyn Samaroo and Jason Wang and Jason Wolfe and Jay Chen and Jay Wang and Jean Harb and Jeffrey Han and Jeffrey Wang and Jennifer Zhao and Jeremy Chen and Jerene Yang and Jerry Tworek and Jesse Chand and Jessica Landon and Jessica Liang and Ji Lin and Jiancheng Liu and Jianfeng Wang and Jie Tang and Jihan Yin and Joanne Jang and Joel Morris and Joey Flynn and Johannes Ferstad and Johannes Heidecke and John Fishbein and John Hallman and Jonah Grant and Jonathan Chien and Jonathan Gordon and Jongsoo Park and Jordan Liss and Jos Kraaijeveld and Joseph Guay and Joseph Mo and Josh Lawson and Josh McGrath and Joshua Vendrow and Joy Jiao and Julian Lee and Julie Steele and Julie Wang and Junhua Mao and Kai Chen and Kai Hayashi and Kai Xiao and Kamyar Salahi and Kan Wu and Karan Sekhri and Karan Sharma and Karan Singhal and Karen Li and Kenny Nguyen and Keren Gu-Lemberg and Kevin King and Kevin Liu and Kevin Stone and Kevin Yu and Kristen Ying and Kristian Georgiev and Kristie Lim and Kushal Tirumala and Kyle Miller and Lama Ahmad and Larry Lv and Laura Clare and Laurance Fauconnet and Lauren Itow and Lauren Yang and Laurentia Romaniuk and Leah Anise and Lee Byron and Leher Pathak and Leon Maksin and Leyan Lo and Leyton Ho and Li Jing and Liang Wu and Liang Xiong and Lien Mamitsuka and Lin Yang and Lindsay McCallum and Lindsey Held and Liz Bourgeois and Logan Engstrom and Lorenz Kuhn and Louis Feuvrier and Lu Zhang and Lucas Switzer and Lukas Kondraciuk and Lukasz Kaiser and Manas Joglekar and Mandeep Singh and Mandip Shah and Manuka Stratta and Marcus Williams and Mark Chen and Mark Sun and Marselus Cayton and Martin Li and Marvin Zhang and Marwan Aljubeh and Matt Nichols and Matthew Haines and Max Schwarzer and Mayank Gupta and Meghan Shah and Melody Huang and Meng Dong and Mengqing Wang and Mia Glaese and Micah Carroll and Michael Lampe and Michael Malek and Michael Sharman and Michael Zhang and Michele Wang and Michelle Pokrass and Mihai Florian and Mikhail Pavlov and Miles Wang and Ming Chen and Mingxuan Wang and Minnia Feng and Mo Bavarian and Molly Lin and Moose Abdool and Mostafa Rohaninejad and Nacho Soto and Natalie Staudacher and Natan LaFontaine and Nathan Marwell and Nelson Liu and Nick Preston and Nick Turley and Nicklas Ansman and Nicole Blades and Nikil Pancha and Nikita Mikhaylin and Niko Felix and Nikunj Handa and Nishant Rai and Nitish Keskar and Noam Brown and Ofir Nachum and Oleg Boiko and Oleg Murk and Olivia Watkins and Oona Gleeson and Pamela Mishkin and Patryk Lesiewicz and Paul Baltescu and Pavel Belov and Peter Zhokhov and Philip Pronin and Phillip Guo and Phoebe Thacker and Qi Liu and Qiming Yuan and Qinghua Liu and Rachel Dias and Rachel Puckett and Rahul Arora and Ravi Teja Mullapudi and Raz Gaon and Reah Miyara and Rennie Song and Rishabh Aggarwal and RJ Marsan and Robel Yemiru and Robert Xiong and Rohan Kshirsagar and Rohan Nuttall and Roman Tsiupa and Ronen Eldan and Rose Wang and Roshan James and Roy Ziv and Rui Shu and Ruslan Nigmatullin and Saachi Jain and Saam Talaie and Sam Altman and Sam Arnesen and Sam Toizer and Sam Toyer and Samuel Miserendino and Sandhini Agarwal and Sarah Yoo and Savannah Heon and Scott Ethersmith and Sean Grove and Sean Taylor and Sebastien Bubeck and Sever Banesiu and Shaokyi Amdo and Shengjia Zhao and Sherwin Wu and Shibani Santurkar and Shiyu Zhao and Shraman Ray Chaudhuri and Shreyas Krishnaswamy and Shuaiqi and Xia and Shuyang Cheng and Shyamal Anadkat and Simón Posada Fishman and Simon Tobin and Siyuan Fu and Somay Jain and Song Mei and Sonya Egoian and Spencer Kim and Spug Golden and SQ Mah and Steph Lin and Stephen Imm and Steve Sharpe and Steve Yadlowsky and Sulman Choudhry and Sungwon Eum and Suvansh Sanjeev and Tabarak Khan and Tal Stramer and Tao Wang and Tao Xin and Tarun Gogineni and Taya Christianson and Ted Sanders and Tejal Patwardhan and Thomas Degry and Thomas Shadwell and Tianfu Fu and Tianshi Gao and Timur Garipov and Tina Sriskandarajah and Toki Sherbakov and Tomer Kaftan and Tomo Hiratsuka and Tongzhou Wang and Tony Song and Tony Zhao and Troy Peterson and Val Kharitonov and Victoria Chernova and Vineet Kosaraju and Vishal Kuo and Vitchyr Pong and Vivek Verma and Vlad Petrov and Wanning Jiang and Weixing Zhang and Wenda Zhou and Wenlei Xie and Wenting Zhan and Wes McCabe and Will DePue and Will Ellsworth and Wulfie Bain and Wyatt Thompson and Xiangning Chen and Xiangyu Qi and Xin Xiang and Xinwei Shi and Yann Dubois and Yaodong Yu and Yara Khakbaz and Yifan Wu and Yilei Qian and Yin Tat Lee and Yinbo Chen and Yizhen Zhang and Yizhong Xiong and Yonglong Tian and Young Cha and Yu Bai and Yu Yang and Yuan Yuan and Yuanzhi Li and Yufeng Zhang and Yuguang Yang and Yujia Jin and Yun Jiang and Yunyun Wang and Yushi Wang and Yutian Liu and Zach Stubenvoll and Zehao Dou and Zheng Wu and Zhigang Wang},
      year={2025},
      eprint={2601.03267},
      archivePrefix={arXiv},
      primaryClass={cs.CL},
      url={https://arxiv.org/abs/2601.03267}, 
}

@misc{openai2024gpt4ocard,
      title={GPT-4o System Card}, 
      author={OpenAI and : and Aaron Hurst and Adam Lerer and Adam P. Goucher and Adam Perelman and Aditya Ramesh and Aidan Clark and AJ Ostrow and Akila Welihinda and Alan Hayes and Alec Radford and Aleksander Mądry and Alex Baker-Whitcomb and Alex Beutel and Alex Borzunov and Alex Carney and Alex Chow and Alex Kirillov and Alex Nichol and Alex Paino and Alex Renzin and Alex Tachard Passos and Alexander Kirillov and Alexi Christakis and Alexis Conneau and Ali Kamali and Allan Jabri and Allison Moyer and Allison Tam and Amadou Crookes and Amin Tootoochian and Amin Tootoonchian and Ananya Kumar and Andrea Vallone and Andrej Karpathy and Andrew Braunstein and Andrew Cann and Andrew Codispoti and Andrew Galu and Andrew Kondrich and Andrew Tulloch and Andrey Mishchenko and Angela Baek and Angela Jiang and Antoine Pelisse and Antonia Woodford and Anuj Gosalia and Arka Dhar and Ashley Pantuliano and Avi Nayak and Avital Oliver and Barret Zoph and Behrooz Ghorbani and Ben Leimberger and Ben Rossen and Ben Sokolowsky and Ben Wang and Benjamin Zweig and Beth Hoover and Blake Samic and Bob McGrew and Bobby Spero and Bogo Giertler and Bowen Cheng and Brad Lightcap and Brandon Walkin and Brendan Quinn and Brian Guarraci and Brian Hsu and Bright Kellogg and Brydon Eastman and Camillo Lugaresi and Carroll Wainwright and Cary Bassin and Cary Hudson and Casey Chu and Chad Nelson and Chak Li and Chan Jun Shern and Channing Conger and Charlotte Barette and Chelsea Voss and Chen Ding and Cheng Lu and Chong Zhang and Chris Beaumont and Chris Hallacy and Chris Koch and Christian Gibson and Christina Kim and Christine Choi and Christine McLeavey and Christopher Hesse and Claudia Fischer and Clemens Winter and Coley Czarnecki and Colin Jarvis and Colin Wei and Constantin Koumouzelis and Dane Sherburn and Daniel Kappler and Daniel Levin and Daniel Levy and David Carr and David Farhi and David Mely and David Robinson and David Sasaki and Denny Jin and Dev Valladares and Dimitris Tsipras and Doug Li and Duc Phong Nguyen and Duncan Findlay and Edede Oiwoh and Edmund Wong and Ehsan Asdar and Elizabeth Proehl and Elizabeth Yang and Eric Antonow and Eric Kramer and Eric Peterson and Eric Sigler and Eric Wallace and Eugene Brevdo and Evan Mays and Farzad Khorasani and Felipe Petroski Such and Filippo Raso and Francis Zhang and Fred von Lohmann and Freddie Sulit and Gabriel Goh and Gene Oden and Geoff Salmon and Giulio Starace and Greg Brockman and Hadi Salman and Haiming Bao and Haitang Hu and Hannah Wong and Haoyu Wang and Heather Schmidt and Heather Whitney and Heewoo Jun and Hendrik Kirchner and Henrique Ponde de Oliveira Pinto and Hongyu Ren and Huiwen Chang and Hyung Won Chung and Ian Kivlichan and Ian O'Connell and Ian O'Connell and Ian Osband and Ian Silber and Ian Sohl and Ibrahim Okuyucu and Ikai Lan and Ilya Kostrikov and Ilya Sutskever and Ingmar Kanitscheider and Ishaan Gulrajani and Jacob Coxon and Jacob Menick and Jakub Pachocki and James Aung and James Betker and James Crooks and James Lennon and Jamie Kiros and Jan Leike and Jane Park and Jason Kwon and Jason Phang and Jason Teplitz and Jason Wei and Jason Wolfe and Jay Chen and Jeff Harris and Jenia Varavva and Jessica Gan Lee and Jessica Shieh and Ji Lin and Jiahui Yu and Jiayi Weng and Jie Tang and Jieqi Yu and Joanne Jang and Joaquin Quinonero Candela and Joe Beutler and Joe Landers and Joel Parish and Johannes Heidecke and John Schulman and Jonathan Lachman and Jonathan McKay and Jonathan Uesato and Jonathan Ward and Jong Wook Kim and Joost Huizinga and Jordan Sitkin and Jos Kraaijeveld and Josh Gross and Josh Kaplan and Josh Snyder and Joshua Achiam and Joy Jiao and Joyce Lee and Juntang Zhuang and Justyn Harriman and Kai Fricke and Kai Hayashi and Karan Singhal and Katy Shi and Kavin Karthik and Kayla Wood and Kendra Rimbach and Kenny Hsu and Kenny Nguyen and Keren Gu-Lemberg and Kevin Button and Kevin Liu and Kiel Howe and Krithika Muthukumar and Kyle Luther and Lama Ahmad and Larry Kai and Lauren Itow and Lauren Workman and Leher Pathak and Leo Chen and Li Jing and Lia Guy and Liam Fedus and Liang Zhou and Lien Mamitsuka and Lilian Weng and Lindsay McCallum and Lindsey Held and Long Ouyang and Louis Feuvrier and Lu Zhang and Lukas Kondraciuk and Lukasz Kaiser and Luke Hewitt and Luke Metz and Lyric Doshi and Mada Aflak and Maddie Simens and Madelaine Boyd and Madeleine Thompson and Marat Dukhan and Mark Chen and Mark Gray and Mark Hudnall and Marvin Zhang and Marwan Aljubeh and Mateusz Litwin and Matthew Zeng and Max Johnson and Maya Shetty and Mayank Gupta and Meghan Shah and Mehmet Yatbaz and Meng Jia Yang and Mengchao Zhong and Mia Glaese and Mianna Chen and Michael Janner and Michael Lampe and Michael Petrov and Michael Wu and Michele Wang and Michelle Fradin and Michelle Pokrass and Miguel Castro and Miguel Oom Temudo de Castro and Mikhail Pavlov and Miles Brundage and Miles Wang and Minal Khan and Mira Murati and Mo Bavarian and Molly Lin and Murat Yesildal and Nacho Soto and Natalia Gimelshein and Natalie Cone and Natalie Staudacher and Natalie Summers and Natan LaFontaine and Neil Chowdhury and Nick Ryder and Nick Stathas and Nick Turley and Nik Tezak and Niko Felix and Nithanth Kudige and Nitish Keskar and Noah Deutsch and Noel Bundick and Nora Puckett and Ofir Nachum and Ola Okelola and Oleg Boiko and Oleg Murk and Oliver Jaffe and Olivia Watkins and Olivier Godement and Owen Campbell-Moore and Patrick Chao and Paul McMillan and Pavel Belov and Peng Su and Peter Bak and Peter Bakkum and Peter Deng and Peter Dolan and Peter Hoeschele and Peter Welinder and Phil Tillet and Philip Pronin and Philippe Tillet and Prafulla Dhariwal and Qiming Yuan and Rachel Dias and Rachel Lim and Rahul Arora and Rajan Troll and Randall Lin and Rapha Gontijo Lopes and Raul Puri and Reah Miyara and Reimar Leike and Renaud Gaubert and Reza Zamani and Ricky Wang and Rob Donnelly and Rob Honsby and Rocky Smith and Rohan Sahai and Rohit Ramchandani and Romain Huet and Rory Carmichael and Rowan Zellers and Roy Chen and Ruby Chen and Ruslan Nigmatullin and Ryan Cheu and Saachi Jain and Sam Altman and Sam Schoenholz and Sam Toizer and Samuel Miserendino and Sandhini Agarwal and Sara Culver and Scott Ethersmith and Scott Gray and Sean Grove and Sean Metzger and Shamez Hermani and Shantanu Jain and Shengjia Zhao and Sherwin Wu and Shino Jomoto and Shirong Wu and Shuaiqi and Xia and Sonia Phene and Spencer Papay and Srinivas Narayanan and Steve Coffey and Steve Lee and Stewart Hall and Suchir Balaji and Tal Broda and Tal Stramer and Tao Xu and Tarun Gogineni and Taya Christianson and Ted Sanders and Tejal Patwardhan and Thomas Cunninghman and Thomas Degry and Thomas Dimson and Thomas Raoux and Thomas Shadwell and Tianhao Zheng and Todd Underwood and Todor Markov and Toki Sherbakov and Tom Rubin and Tom Stasi and Tomer Kaftan and Tristan Heywood and Troy Peterson and Tyce Walters and Tyna Eloundou and Valerie Qi and Veit Moeller and Vinnie Monaco and Vishal Kuo and Vlad Fomenko and Wayne Chang and Weiyi Zheng and Wenda Zhou and Wesam Manassra and Will Sheu and Wojciech Zaremba and Yash Patil and Yilei Qian and Yongjik Kim and Youlong Cheng and Yu Zhang and Yuchen He and Yuchen Zhang and Yujia Jin and Yunxing Dai and Yury Malkov},
      year={2024},
      eprint={2410.21276},
      archivePrefix={arXiv},
      primaryClass={cs.CL},
      url={https://arxiv.org/abs/2410.21276}, 
}

@misc{monroe2015learningrationalspeechacts,
      title={Learning in the Rational Speech Acts Model}, 
      author={Will Monroe and Christopher Potts},
      year={2015},
      eprint={1510.06807},
      archivePrefix={arXiv},
      primaryClass={cs.CL},
      url={https://arxiv.org/abs/1510.06807}, 
}

@misc{zhixuan2024pragmaticinstructionfollowinggoal,
      title={Pragmatic Instruction Following and Goal Assistance via Cooperative Language-Guided Inverse Planning}, 
      author={Tan Zhi-Xuan and Lance Ying and Vikash Mansinghka and Joshua B. Tenenbaum},
      year={2024},
      eprint={2402.17930},
      archivePrefix={arXiv},
      primaryClass={cs.AI},
      url={https://arxiv.org/abs/2402.17930}, 
}

@misc{wang2026emotionthinkerprosodyawarereinforcementlearning,
      title={EmotionThinker: Prosody-Aware Reinforcement Learning for Explainable Speech Emotion Reasoning}, 
      author={Dingdong Wang and Shujie Liu and Tianhua Zhang and Youjun Chen and Jinyu Li and Helen Meng},
      year={2026},
      eprint={2601.15668},
      archivePrefix={arXiv},
      primaryClass={cs.SD},
      url={https://arxiv.org/abs/2601.15668}, 
}

@misc{xu2023secapspeechemotioncaptioning,
      title={SECap: Speech Emotion Captioning with Large Language Model}, 
      author={Yaoxun Xu and Hangting Chen and Jianwei Yu and Qiaochu Huang and Zhiyong Wu and Shixiong Zhang and Guangzhi Li and Yi Luo and Rongzhi Gu},
      year={2023},
      eprint={2312.10381},
      archivePrefix={arXiv},
      primaryClass={cs.SD},
      url={https://arxiv.org/abs/2312.10381}, 
}

@misc{zhang2025classificationspeechemotionreasoning,
      title={Beyond Classification: Towards Speech Emotion Reasoning with Multitask AudioLLMs}, 
      author={Wenyu Zhang and Yingxu He and Geyu Lin and Zhuohan Liu and Shuo Sun and Bin Wang and Xunlong Zou and Jeremy H. M. Wong and Qiongqiong Wang and Hardik B. Sailor and Nancy F. Chen and Ai Ti Aw},
      year={2025},
      eprint={2506.06820},
      archivePrefix={arXiv},
      primaryClass={cs.CL},
      url={https://arxiv.org/abs/2506.06820}, 
}

@misc{arxiv:2503.04721,
  title = {Full-Duplex-Bench: A Benchmark to Evaluate Full-duplex Spoken Dialogue Models on Turn-taking Capabilities},
  author = {Guan-Ting Lin and Jiachen Lian and Tingle Li and Qirui Wang and Gopala Anumanchipalli and Alexander H. Liu and Hung-yi Lee},
  year = {2025},
  eprint = {2503.04721},
  archivePrefix = {arXiv},
  primaryClass = {cs.CL},
  url = {https://arxiv.org/abs/2503.04721}
}

@misc{arxiv:2203.16502,
  title = {Generative Spoken Dialogue Language Modeling},
  author = {Tu Anh Nguyen and Eugene Kharitonov and Jade Copet and Yossi Adi and Wei-Ning Hsu and Ali Elkahky and Paden Tomasello and Robin Algayres and Beno{\^\i}t Sagot and Abdelrahman Mohamed and Emmanuel Dupoux},
  year = {2022},
  eprint = {2203.16502},
  archivePrefix = {arXiv},
  primaryClass = {cs.CL},
  url = {https://arxiv.org/abs/2203.16502}
}

@misc{arxiv:2503.01174,
  title = {Talking Turns: Benchmarking Audio Foundation Models on Turn-Taking Dynamics},
  author = {Siddhant Arora and Zhiyun Lu and Chung-Cheng Chiu and Ruoming Pang and Shinji Watanabe},
  year = {2025},
  eprint = {2503.01174},
  archivePrefix = {arXiv},
  primaryClass = {cs.CL},
  url = {https://arxiv.org/abs/2503.01174}
}

@article{gravano2011turn,
  title={Turn-taking cues in task-oriented dialogue},
  author={Gravano, Agust{\'\i}n and Hirschberg, Julia},
  journal={Computer Speech \& Language},
  volume={25},
  number={3},
  pages={601--634},
  year={2011},
  publisher={Elsevier}
}

@article{duncan1972some,
  title={Some signals and rules for taking speaking turns in conversations.},
  author={Duncan, Starkey},
  journal={Journal of personality and social psychology},
  volume={23},
  number={2},
  pages={283},
  year={1972},
  publisher={American Psychological Association}
}

@article{raux2012optimizing,
  title={Optimizing the turn-taking behavior of task-oriented spoken dialog systems},
  author={Raux, Antoine and Eskenazi, Maxine},
  journal={ACM Transactions on Speech and Language Processing (TSLP)},
  volume={9},
  number={1},
  pages={1--23},
  year={2012},
  publisher={ACM New York, NY, USA}
}

@article{schegloff1982discourse,
  title={Discourse as an interactional achievement: Some uses of ‘uh huh’and other things that come between sentences},
  author={Schegloff, Emanuel A},
  journal={Analyzing discourse: Text and talk},
  volume={71},
  number={93},
  year={1982}
}

@inproceedings{khouzaimi2016reinforcement,
  title={Reinforcement Learning for Turn-Taking Management in Incremental Spoken Dialogue Systems.},
  author={Khouzaimi, Hatim and Laroche, Romain and Lef{\`e}vre, Fabrice},
  booktitle={IJCAI},
  pages={2831--2837},
  year={2016}
}

@article{marge2022spoken,
  title={Spoken language interaction with robots: Recommendations for future research},
  author={Marge, Matthew and Espy-Wilson, Carol and Ward, Nigel G and Alwan, Abeer and Artzi, Yoav and Bansal, Mohit and Blankenship, Gil and Chai, Joyce and Daum{\'e} III, Hal and Dey, Debadeepta and others},
  journal={Computer Speech \& Language},
  volume={71},
  pages={101255},
  year={2022},
  publisher={Elsevier}
}

@article{du2024cosyvoice,
  title={Cosyvoice 2: Scalable streaming speech synthesis with large language models},
  author={Du, Zhihao and Wang, Yuxuan and Chen, Qian and Shi, Xian and Lv, Xiang and Zhao, Tianyu and Gao, Zhifu and Yang, Yexin and Gao, Changfeng and Wang, Hui and others},
  journal={arXiv preprint arXiv:2412.10117},
  year={2024}
}

@article{reece2023candor,
  title={The CANDOR corpus: Insights from a large multimodal dataset of naturalistic conversation},
  author={Reece, Andrew and Cooney, Gus and Bull, Peter and Chung, Christine and Dawson, Bryn and Fitzpatrick, Casey and Glazer, Tamara and Knox, Dean and Liebscher, Alex and Marin, Sebastian},
  journal={Science Advances},
  volume={9},
  number={13},
  pages={eadf3197},
  year={2023},
  publisher={American Association for the Advancement of Science}
}

@misc{got,
      title={Beyond Chain-of-Thought, Effective Graph-of-Thought Reasoning in Language Models}, 
      author={Yao Yao and Zuchao Li and Hai Zhao},
      year={2024},
      eprint={2305.16582},
      archivePrefix={arXiv},
      primaryClass={cs.CL},
      url={https://arxiv.org/abs/2305.16582}, 
}

@inproceedings{panayotov2015librispeech,
  title={Librispeech: an asr corpus based on public domain audio books},
  author={Panayotov, Vassil and Chen, Guoguo and Povey, Daniel and Khudanpur, Sanjeev},
  booktitle={2015 IEEE international conference on acoustics, speech and signal processing (ICASSP)},
  pages={5206--5210},
  year={2015},
  organization={IEEE}
}

@inproceedings{hara18_interspeech,
  title     = {Prediction of Turn-taking Using Multitask Learning with Prediction of Backchannels and Fillers},
  author    = {Kohei Hara and Koji Inoue and Katsuya Takanashi and Tatsuya Kawahara},
  year      = {2018},
  booktitle = {Interspeech 2018},
  pages     = {991--995},
  doi       = {10.21437/Interspeech.2018-1442},
  issn      = {2958-1796},
}

@inproceedings{li-etal-2022-speak,
    title = "When can {I} Speak? Predicting initiation points for spoken dialogue agents",
    author = "Li, Siyan  and
      Paranjape, Ashwin  and
      Manning, Christopher",
    editor = "Lemon, Oliver  and
      Hakkani-Tur, Dilek  and
      Li, Junyi Jessy  and
      Ashrafzadeh, Arash  and
      Garcia, Daniel Hern{\'a}ndez  and
      Alikhani, Malihe  and
      Vandyke, David  and
      Du{\v{s}}ek, Ond{\v{r}}ej",
    booktitle = "Proceedings of the 23rd Annual Meeting of the Special Interest Group on Discourse and Dialogue",
    month = sep,
    year = "2022",
    address = "Edinburgh, UK",
    publisher = "Association for Computational Linguistics",
    url = "https://aclanthology.org/2022.sigdial-1.22/",
    doi = "10.18653/v1/2022.sigdial-1.22",
    pages = "217--224",

}

@book{jurafsky2025speech,
  title        = {Speech and Language Processing},
  edition      = {3},
  author       = {Jurafsky, Daniel and Martin, James H.},
  year         = {2025},
  publisher    = {Draft / Stanford University},
  chapter      = {15},
  url          = {https://web.stanford.edu/~jurafsky/slp3/15.pdf}
}

@INPROCEEDINGS{5494991,
  author={Lee, Chi-Chun and Narayanan, Shrikanth},
  booktitle={2010 IEEE International Conference on Acoustics, Speech and Signal Processing}, 
  title={Predicting interruptions in dyadic spoken interactions}, 
  year={2010},
  volume={},
  number={},
  pages={5250-5253},
 }

@misc{vap,
      title={Yeah, Un, Oh: Continuous and Real-time Backchannel Prediction with Fine-tuning of Voice Activity Projection}, 
      author={Koji Inoue and Divesh Lala and Gabriel Skantze and Tatsuya Kawahara},
      year={2025},
      eprint={2410.15929},
      archivePrefix={arXiv},
      primaryClass={cs.CL},
      url={https://arxiv.org/abs/2410.15929}, 
}

@misc{dgslm,
      title={Generative Spoken Dialogue Language Modeling}, 
      author={Tu Anh Nguyen and Eugene Kharitonov and Jade Copet and Yossi Adi and Wei-Ning Hsu and Ali Elkahky and Paden Tomasello and Robin Algayres and Benoit Sagot and Abdelrahman Mohamed and Emmanuel Dupoux},
      year={2022},
      eprint={2203.16502},
      archivePrefix={arXiv},
      primaryClass={cs.CL},
      url={https://arxiv.org/abs/2203.16502}, 
}

@article{defossez2024moshi,
  title={Moshi: a speech-text foundation model for real-time dialogue},
  author={D{\'e}fossez, Alexandre and Mazar{\'e}, Laurent and Orsini, Manu and Royer, Am{\'e}lie and P{\'e}rez, Patrick and J{\'e}gou, Herv{\'e} and Grave, Edouard and Zeghidour, Neil},
  journal={arXiv preprint arXiv:2410.00037},
  year={2024}
}

@inproceedings{inproceedings,
author = {Trinh, Anh Duong and Ross, Robert and Kelleher, John},
year = {2018},
month = {11},
pages = {},
title = {A Multi-Task Approach to Incremental Dialogue State Tracking},
booktitle = {Proceedings of the SemDial Workshop on the Semantics and Pragmatics of Dialogue}
}

@misc{kennington2025priorlessonsincrementaldialogue,
      title={Prior Lessons of Incremental Dialogue and Robot Action Management for the Age of Language Models}, 
      author={Casey Kennington and Pierre Lison and David Schlangen},
      year={2025},
      eprint={2501.00953},
      archivePrefix={arXiv},
      primaryClass={cs.CL},
      url={https://arxiv.org/abs/2501.00953}, 
}

@misc{lee2021dialoguestatetrackinglanguage,
      title={Dialogue State Tracking with a Language Model using Schema-Driven Prompting}, 
      author={Chia-Hsuan Lee and Hao Cheng and Mari Ostendorf},
      year={2021},
      eprint={2109.07506},
      archivePrefix={arXiv},
      primaryClass={cs.CL},
      url={https://arxiv.org/abs/2109.07506}, 
}

@inproceedings{44018,title	= {Machine Learning for Dialog State Tracking: A Review},author	= {Matthew Henderson},year	= {2015},booktitle	= {Proceedings of The First International Workshop on Machine Learning in Spoken Language Processing}}

@inproceedings{metallinou-etal-2013-discriminative,
    title = "Discriminative state tracking for spoken dialog systems",
    author = "Metallinou, Angeliki  and
      Bohus, Dan  and
      Williams, Jason",
    editor = "Schuetze, Hinrich  and
      Fung, Pascale  and
      Poesio, Massimo",
    booktitle = "Proceedings of the 51st Annual Meeting of the Association for Computational Linguistics (Volume 1: Long Papers)",
    month = aug,
    year = "2013",
    address = "Sofia, Bulgaria",
    publisher = "Association for Computational Linguistics",
    url = "https://aclanthology.org/P13-1046/",
    pages = "466--475"
}

@article{williams2012challenges,
author = {Williams, Jason},
title = {Challenges and Opportunities for State Tracking in Statistical Spoken Dialog Systems: Results From Two Public Deployments},
year = {2012},
month = {December},
publisher = {IEEE SPS},
url = {https://www.microsoft.com/en-us/research/publication/challenges-and-opportunities-for-state-tracking-in-statistical-spoken-dialog-systems-results-from-two-public-deployments/},
pages = {959-970},
journal = {IEEE Journal of Selected Topics in Signal Processing},
volume = {6},
edition = {IEEE Journal of Selected Topics in Signal Processing},
}

@misc{potdar2021streamingendtoendframeworkspoken,
  title = {A Streaming End-to-End Framework For Spoken Language Understanding},
  author = {Potdar, Nihal and Avila, Anderson R. and Xing, Chao and Wang, Dong and Cao, Yiran and Chen, Xiao},
  year = {2021},
  eprint = {2105.10042},
  archivePrefix = {arXiv},
  primaryClass = {cs.CL},
  url = {https://arxiv.org/abs/2105.10042}
}

@inproceedings{miah-etal-2023-hierarchical,
    title = "Hierarchical Fusion for Online Multimodal Dialog Act Classification",
    author = "Miah, Md Messal Monem  and
      Pyarelal, Adarsh  and
      Huang, Ruihong",
    editor = "Bouamor, Houda  and
      Pino, Juan  and
      Bali, Kalika",
    booktitle = "Findings of the Association for Computational Linguistics: EMNLP 2023",
    month = dec,
    year = "2023",
    address = "Singapore",
    publisher = "Association for Computational Linguistics",
    url = "https://aclanthology.org/2023.findings-emnlp.505/",
    doi = "10.18653/v1/2023.findings-emnlp.505",
    pages = "7532--7545"
}

@misc{arora2025talkingturnsbenchmarkingaudio,
      title={Talking Turns: Benchmarking Audio Foundation Models on Turn-Taking Dynamics}, 
      author={Siddhant Arora and Zhiyun Lu and Chung-Cheng Chiu and Ruoming Pang and Shinji Watanabe},
      year={2025},
      eprint={2503.01174},
      archivePrefix={arXiv},
      primaryClass={cs.CL},
      url={https://arxiv.org/abs/2503.01174}, 
}

@misc{cao2021sequential,
      title={Sequential End-to-End Intent and Slot Label Classification and Localization}, 
      author={Yiran Cao and Nihal Potdar and Anderson R. Avila},
      year={2021},
      eprint={2106.04660},
      archivePrefix={arXiv},
      primaryClass={cs.CL},
      url={https://arxiv.org/abs/2106.04660}, 
}

@misc{fu2022multitask,
      title={Multi-task RNN-T with Semantic Decoder for Streamable Spoken Language Understanding}, 
      author={Xuandi Fu and Feng-Ju Chang and Martin Radfar and Kai Wei and Jing Liu and Grant P. Strimel and Kanthashree Mysore Sathyendra},
      year={2022},
      eprint={2204.00558},
      archivePrefix={arXiv},
      primaryClass={cs.CL},
      url={https://arxiv.org/abs/2204.00558}, 
}

@misc{arora2022twopass,
      title={Two-Pass Low Latency End-to-End Spoken Language Understanding}, 
      author={Siddhant Arora and Siddharth Dalmia and Xuankai Chang and Brian Yan and Alan Black and Shinji Watanabe},
      year={2022},
      eprint={2207.06670},
      archivePrefix={arXiv},
      primaryClass={cs.CL},
      url={https://arxiv.org/abs/2207.06670}, 
}

@misc{raheja2019dialogue,
      title={Dialogue Act Classification with Context-Aware Self-Attention}, 
      author={Vipul Raheja and Joel Tetreault},
      year={2019},
      eprint={1904.02594},
      archivePrefix={arXiv},
      primaryClass={cs.CL},
      url={https://arxiv.org/abs/1904.02594}, 
}

@misc{he2021speaker,
      title={Speaker Turn Modeling for Dialogue Act Classification}, 
      author={Zihao He and Leili Tavabi and Kristina Lerman and Mohammad Soleymani},
      year={2021},
      eprint={2109.05056},
      archivePrefix={arXiv},
      primaryClass={cs.CL},
      url={https://arxiv.org/abs/2109.05056}, 
}

@misc{chang2022turntaking,
      title={Turn-Taking Prediction for Natural Conversational Speech}, 
      author={Shuo-yiin Chang and Bo Li and Tara N. Sainath and Chao Zhang and Trevor Strohman and Qiao Liang and Yanzhang He},
      year={2022},
      eprint={2208.13321},
      archivePrefix={arXiv},
      primaryClass={cs.CL},
      url={https://arxiv.org/abs/2208.13321}, 
}

@misc{ekstedt2022voice,
      title={Voice Activity Projection: Self-supervised Learning of Turn-taking Events}, 
      author={Erik Ekstedt and Gabriel Skantze},
      year={2022},
      eprint={2205.09812},
      archivePrefix={arXiv},
      primaryClass={eess.AS},
      url={https://arxiv.org/abs/2205.09812}, 
}

@inproceedings{kurata2023multimodal,
  title     = {{Multimodal Turn-Taking Model Using Visual Cues for End-of-Utterance Prediction in Spoken Dialogue Systems}},
  author    = {Fuma Kurata and Mao Saeki and Shinya Fujie and Yoichi Matsuyama},
  year      = {2023},
  booktitle = {{Interspeech 2023}},
  pages     = {2658--2662},
  doi       = {10.21437/Interspeech.2023-578},
  issn      = {2958-1796},
}

@article{Sacks1974TurnTaking, title={A simplest systematics for the organization of turn-taking for conversation}, volume={50}, DOI={10.2307/412243}, number={4}, journal={Language}, author={Sacks, Harvey and Schegloff, Emanuel A. and Jefferson, Gail}, year={1974}, pages={696--735}}

@article{Stivers2009TurnTaking,

  title = {Universals and cultural variation in turn-taking in conversation},

  author = {Stivers, T. and Enfield, N. J. and Brown, P. and Englert, C. and Hayashi, M. and Heinemann, T. and Hoymann, G. and Rossano, F. and de, Ruiter JP and Yoon, K. E. and Levinson, S. C.},

  journal = {Proceedings of the National Academy of Sciences of the United States of America},

  year = {2009},

  volume = {106},

  number = {26},

  pages = {10587--92},

  doi = {10.1073/pnas.0903616106},

  pmid = {19553212},

  url = {https://pubmed.ncbi.nlm.nih.gov/19553212/}

}

@ARTICLE{Levinson2015Timing,
    
AUTHOR={Levinson, Stephen C.  and Torreira, Francisco },
           
TITLE={Timing in turn-taking and its implications for processing models of language},
          
JOURNAL={Frontiers in Psychology},
          
VOLUME={Volume 6 - 2015},
  
YEAR={2015},
  
URL={https://www.frontiersin.org/journals/psychology/articles/10.3389/fpsyg.2015.00731},
  
DOI={10.3389/fpsyg.2015.00731},
  
ISSN={1664-1078}}

@article{Stolcke2000DialogueAct,
   title={Dialogue Act Modeling for Automatic Tagging and Recognition of Conversational Speech},
   volume={26},
   ISSN={1530-9312},
   url={http://dx.doi.org/10.1162/089120100561737},
   DOI={10.1162/089120100561737},
   number={3},
   journal={Computational Linguistics},
   publisher={MIT Press - Journals},
   author={Stolcke, Andreas and Ries, Klaus and Coccaro, Noah and Shriberg, Elizabeth and Bates, Rebecca and Jurafsky, Daniel and Taylor, Paul and Martin, Rachel and Ess-Dykema, Carol Van and Meteer, Marie},
   year={2000},
   month=sep, pages={339--373} }

@misc{ISO24617-2,

  title={ISO 24617-2:2020 Language resource management -- Semantic annotation framework -- Part 2: Dialogue acts},

  author={{International Organization for Standardization}},

  organization={International Organization for Standardization},

  year={2020},

  url={https://www.iso.org/obp/ui/}

}

@inproceedings{Bunt2020ISO,
    title = "The {ISO} Standard for Dialogue Act Annotation, Second Edition",
    author = "Bunt, Harry  and
      Petukhova, Volha  and
      Gilmartin, Emer  and
      Pelachaud, Catherine  and
      Fang, Alex  and
      Keizer, Simon  and
      Pr{\'e}vot, Laurent",
    editor = "Calzolari, Nicoletta  and
      B{\'e}chet, Fr{\'e}d{\'e}ric  and
      Blache, Philippe  and
      Choukri, Khalid  and
      Cieri, Christopher  and
      Declerck, Thierry  and
      Goggi, Sara  and
      Isahara, Hitoshi  and
      Maegaard, Bente  and
      Mariani, Joseph  and
      Mazo, H{\'e}l{\`e}ne  and
      Moreno, Asuncion  and
      Odijk, Jan  and
      Piperidis, Stelios",
    booktitle = "Proceedings of the Twelfth Language Resources and Evaluation Conference",
    month = may,
    year = "2020",
    address = "Marseille, France",
    publisher = "European Language Resources Association",
    url = "https://aclanthology.org/2020.lrec-1.69/",
    pages = "549--558",
    language = "eng",
    ISBN = "979-10-95546-34-4"
}

@inproceedings{Zaidan2007Rationales,
    title = "Using ``Annotator Rationales'' to Improve Machine Learning for Text Categorization",
    author = "Zaidan, Omar  and
      Eisner, Jason  and
      Piatko, Christine",
    editor = "Sidner, Candace  and
      Schultz, Tanja  and
      Stone, Matthew  and
      Zhai, ChengXiang",
    booktitle = "Human Language Technologies 2007: The Conference of the North {A}merican Chapter of the Association for Computational Linguistics; Proceedings of the Main Conference",
    month = apr,
    year = "2007",
    address = "Rochester, New York",
    publisher = "Association for Computational Linguistics",
    url = "https://aclanthology.org/N07-1033/",
    pages = "260--267"
}

@misc{DeYoung2020ERASER,
      title={ERASER: A Benchmark to Evaluate Rationalized NLP Models}, 
      author={Jay DeYoung and Sarthak Jain and Nazneen Fatema Rajani and Eric Lehman and Caiming Xiong and Richard Socher and Byron C. Wallace},
      year={2020},
      eprint={1911.03429},
      archivePrefix={arXiv},
      primaryClass={cs.CL},
      url={https://arxiv.org/abs/1911.03429}, 
}

@inproceedings{Thompson1993MapTask, 
    title = "The {HCRC} Map Task Corpus: Natural Dialogue for Speech Recognition",
    author = "Thompson, Henry S.  and
      Anderson, Anne  and
      Bard, Ellen Gurman  and
      Doherty-Sneddon, Gwyneth  and
      Newlands, Alison  and
      Sotillo, Cathy",
    booktitle = "{H}uman {L}anguage {T}echnology: Proceedings of a Workshop Held at Plainsboro, New Jersey, March 21-24, 1993",
    year = "1993",
    url = "https://aclanthology.org/H93-1005/"
}

@inproceedings{Chen2024RASwDA,

  title = {{RASwDA}: Re-Aligned Switchboard Dialog Act Corpus for Dialog Act Modeling in Conversations},

  author = {Chen, Ruizhe and Lin, Zhouhan and Hirschberg, Julia},

  booktitle = {Proceedings of the 15th International Workshop on Spoken Dialogue Systems Technology},

  year = {2024},

  url = {https://www.cs.columbia.edu/speech/PaperFiles/2024/iwsds24_raswda_paper.pdf}

}

@inproceedings{Shriberg2004MRDA,
    title = "The {ICSI} Meeting Recorder Dialog Act ({MRDA}) Corpus",
    author = "Shriberg, Elizabeth  and
      Dhillon, Raj  and
      Bhagat, Sonali  and
      Ang, Jeremy  and
      Carvey, Hannah",
    booktitle = "Proceedings of the 5th {SIG}dial Workshop on Discourse and Dialogue at {HLT}-{NAACL} 2004",
    month = apr # " 30 - " # may # " 1",
    year = "2004",
    address = "Cambridge, Massachusetts, USA",
    publisher = "Association for Computational Linguistics",
    url = "https://aclanthology.org/W04-2319/",
    pages = "97--100"
}

@misc{arora2025talkingturns,
      title={Talking Turns: Benchmarking Audio Foundation Models on Turn-Taking Dynamics}, 
      author={Siddhant Arora and Zhiyun Lu and Chung-Cheng Chiu and Ruoming Pang and Shinji Watanabe},
      year={2025},
      eprint={2503.01174},
      archivePrefix={arXiv},
      primaryClass={cs.CL},
      url={https://arxiv.org/abs/2503.01174}, 
}

@misc{yu2021midas,
      title={MIDAS: A Dialog Act Annotation Scheme for Open Domain Human Machine Spoken Conversations}, 
      author={Dian Yu and Zhou Yu},
      year={2019},
      eprint={1908.10023},
      archivePrefix={arXiv},
      primaryClass={cs.CL},
      url={https://arxiv.org/abs/1908.10023}, 
}

\appendix

\section{Appendix}
\section{ConversationGoT-120h}

\subsection{Dataset Construction}

\paragraph{Dialogue Text Generation.}
ConversationGoT-120h targets open-domain chit-chat dialogues with diverse topics, natural topic transitions, and rich character settings. We design a structured generation pipeline to construct the dialogue text. First, we use GPT-4o~\citep{openai2024gpt4ocard} to synthesize speaker identity information, including interests, birthplace, educational background, and family background. Based on the generated identities, we then produce 8--10 topics for each dialogue and divide them into interactive topics, which are jointly discussed by both speakers, and expressive topics, which are primarily led by a single speaker. During topic-level generation, we mask the full identity profile and condition each generation step on only 1--2 identity traits, which reduces repetition and improves content diversity. Finally, given the identity information and topic sequence, we generate topic transitions, the dialogue beginning, and the dialogue ending to improve global coherence.

\paragraph{Causal Annotation Generation.}
To provide supervision signals consistent with a strictly causal setting, we model each dialogue at a one-second resolution. We segment the audio and text into consecutive 1-second chunks and generate hierarchical conversational behavior states and rationale annotations without relying on the global future dialogue structure. Specifically, for each second, we first use GPT-4o~\citep{openai2024gpt4ocard} to retrieve remote content-anchor sentences relevant to the current behavior state and organize them from far to near into a topic thought chain. We then feed this topic chain, together with the observable inputs at the current second, into GPT-5~\citep{singh2025openaigpt5card}, which derives the current conversational behavior state, including the high-level communicative function and the low-level interaction behavior. The step-by-step reasoning process is saved as the rationale annotation. Meanwhile, the selected anchors provide multi-hot supervision for Stage-1 evidence selection. In this way, the dataset supervises GoT to learn evidence-grounded reasoning chains for streaming decision making, rather than only providing post-hoc explanations for predefined labels.

\paragraph{Speech Synthesis.}
Finally, we synthesize speech for the generated dialogues. We collect 1,166 high-quality reference voices from LibriSpeech~\citep{panayotov2015librispeech} and use CosyVoice2~\citep{du2024cosyvoice} to generate speech with diverse prosody and conversational speaking styles. The resulting dataset contains 720 samples, each with an average duration of approximately 5 minutes, totaling around 60 hours of audio.

\subsection{Conversational Behavior State Definition}

A conversational behavior state consists of a low-level interaction behavior and a high-level communicative function. All conversational behavior states are generated under strictly causal inputs. At time $t$, the annotation model only accesses information available up to $t$, ensuring that no future events or external information leak into the current behavior-state decision.

\paragraph{Low-Level Interaction Behaviors.}
\begin{itemize}
    \item \textbf{Silence}: no active speech is produced by either speaker during the current time segment.
    \item \textbf{Backchannel}: the speaker changes, but turn ownership does not change, such as a brief acknowledgment or agreement followed by a return to the prior speaker or topic.
    \item \textbf{Interruption}: the speaker changes and turn ownership changes rapidly within a short time, often involving overlap or both speakers attempting to take the turn.
    \item \textbf{Turn-taking}: the speaker changes and turn ownership changes normally, corresponding to a smooth turn transition.
    \item \textbf{Continuation}: the speaker does not change and continues the current turn.
\end{itemize}

\paragraph{High-Level Communicative Functions.}
\begin{itemize}
    \item \textbf{Constatives}: the intent to state or describe facts, opinions, or information.
    \item \textbf{Directives}: the intent to make a request, ask a question, give an instruction, or guide the other party to act.
    \item \textbf{Acknowledgments}: the intent to confirm, agree with, thank, apologize for, or otherwise maintain interaction regarding the other party's content.
    \item \textbf{Commissives}: the intent to express a commitment, plan, willingness, or future action.
\end{itemize}

\subsection{Additional Dataset Diagnostics}

\paragraph{Overlap and Interaction Quality.}
Because ConversationGoT-120h targets full-duplex conversational behavior, we examine whether the simulated dialogues contain sufficient overlap and turn-level interaction events. Table~\ref{tab:quality-check1} reports turn-taking event frequencies and cumulative durations for our simulation data, a human reference, and model baselines.

\begin{table*}[t]
\centering
\scriptsize
\caption{Turn-taking event frequencies per minute and cumulative durations for the simulation dataset, a human reference, and model baselines. Human, dGSLM, and Moshi values are reproduced from Fig.~2 of~\citep{arxiv:2503.01174}.}
\label{tab:quality-check1}
\begin{tabular}{lcccccccc}
\toprule
& \multicolumn{4}{c}{\textit{Number of events per minute}} 
& \multicolumn{4}{c}{\textit{Cumulative duration (\% of time)}} \\
\textbf{Event type} 
& \textbf{Simulation} & \textbf{Human} & \textbf{dGSLM} & \textbf{Moshi} 
& \textbf{Simulation} & \textbf{Human} & \textbf{dGSLM} & \textbf{Moshi} \\
\midrule
IPU       & 23.06 & 15.7  & 24.2  & 21.6  & 84.7 & 97.3 & 99.0 & 81.0 \\
Pause     & 10.7  & 3.8   & 5.4   & 10.2  & 9.6  & 5.7  & 6.0  & 10.3 \\
Gap       & 7.3   & 5.5   & 7.2   & 6.7   & 1.6  & 3.7  & 4.8  & 11.8 \\
Overlap   & 6.7   & 6.6   & 10.9  & 4.8   & 4.2  & 6.7  & 9.7  & 3.1 \\
\bottomrule
\end{tabular}
\end{table*}

As shown in Table~\ref{tab:quality-check1}, our simulation data exhibit denser micro-segmentation than human dialogues, with higher IPU and pause counts per minute. At the same time, the gap and overlap frequencies are close to the human reference. The cumulative-duration statistics indicate shorter overlaps and more within-speaker pauses, suggesting that the simulated dialogues contain frequent short backchannels, hesitations, and local interaction dynamics rather than only long monologic turns. Overall, these statistics suggest that ConversationGoT-120h preserves key interactional properties of full-duplex conversations, while still differing from human conversations in micro-segmentation density.

\paragraph{Event Distribution and Anchor Statistics.}
ConversationGoT-120h shows clear head concentration in both high-level and low-level speech-act dimensions. Among the four high-level classes, Constatives account for $54.18\%$ of the labels, followed by Directives ($18.93\%$), Acknowledgments ($14.43\%$), and Commissives ($12.37\%$). This distribution indicates that the corpus is centered on statement and description functions, while other communicative functions are more dispersed.

The imbalance is stronger for the five low-level interaction behaviors. Continuation accounts for $64.77\%$ of the labels, followed by Turn-taking ($19.03\%$), while Interruption ($9.07\%$) and Backchannel ($7.13\%$) form the long tail. This distribution reflects the fact that dialogue progression mainly consists of continuous speaking and regular turn transitions, whereas interruption and feedback signals occur less frequently.

For anchor supervision, each segment contains an average of $3.92$ anchors with a standard deviation of $1.72$, and the 95th percentile is $7.48$ anchors. The average anchor distance is $60.55$ seconds, indicating that selected evidence often comes from non-local dialogue history rather than only adjacent context.

\subsection{Human Evaluation Protocol}

\paragraph{Data Verification Protocol.}
To ensure the quality and usability of the synthesized data, all generated samples undergo human review. If a sample's score falls below 70\% of the maximum score defined in the rating rubric, the reviewer discards the sample and triggers the pipeline to regenerate a replacement. This process is repeated until a qualified sample is obtained. Failed samples are retained as case studies to support iterative improvements to the generation and annotation pipeline.

\paragraph{Human-Model Agreement.}
We use Human-Model Agreement (HMA) to measure whether human volunteers agree with model predictions at each second. For high-level and low-level labels, HMA is defined as:
$\mathrm{HMA}^{h}
=
\frac{1}{TR}\sum_{t=1}^{T}\sum_{r=1}^{R} a^{h}_{t,r}$.

$\mathrm{HMA}^{l}
=
\frac{1}{TR}\sum_{t=1}^{T}\sum_{r=1}^{R} a^{l}_{t,r}$.
Here, $T$ is the number of evaluated seconds, $R$ is the number of volunteers, and $a^{h}_{t,r}, a^{l}_{t,r}\in\{0,1\}$ indicate whether volunteer $r$ agrees with the model's decision at second $t$ for the high-level and low-level labels, respectively.

\paragraph{Dialogue Quality Rubric.}
We ask human raters to evaluate dialogue quality along four dimensions.

\paragraph{Dimension 1: Naturalness \& Flow.}
\textbf{Question:} Does the dialogue read naturally? Do turns and topics flow smoothly?
\begin{itemize}
  \item \textbf{9--10: Very natural.}
  Strong colloquial feel with almost no awkward phrasing; responses clearly build on the previous turn; topic transitions are explicit and smooth, making the dialogue feel like a real conversation.
  \item \textbf{7--8: Mostly natural.}
  Most sentences are fluent; a few feel slightly stiff; turn-to-turn flow is generally coherent; topic transitions may be slightly abrupt but remain acceptable.
  \item \textbf{5--6: Fair.}
  Understandable but somewhat written or templated; some loose connections or mild topic drifting; several topic shifts are obvious hard cuts.
  \item \textbf{3--4: Poor.}
  Many sentences feel unnatural for spoken language; turns often fail to respond to each other; topic jumps feel random or poorly motivated.
  \item \textbf{1--2: Very poor.}
  Heavy machine-translation or stitched-together feel; dialogue is largely disjointed and not conversational; topic changes feel completely random.
\end{itemize}

\paragraph{Dimension 2: Logical Consistency \& Identity Coherence.}
\textbf{Question:} Is the dialogue logically consistent and aligned with the given identities?
\begin{itemize}
  \item \textbf{9--10: Very consistent.}
  No obvious contradictions; frequently and correctly reuses earlier details such as hometown, major, or family background; topics strongly reflect the characters' identities.
  \item \textbf{7--8: Mostly consistent.}
  Overall logic makes sense with only minor fuzziness; most content fits the identities; a few points may be slightly stretched but still plausible.
  \item \textbf{5--6: Fair.}
  Generally coherent, but includes a few inconsistencies or forced elements; identity signals are weak and the dialogue could almost be said by anyone.
  \item \textbf{3--4: Poor.}
  Noticeable conflicts with earlier facts, such as places, jobs, family, or timeline; topics often feel unrelated to the character setup.
  \item \textbf{1--2: Very poor.}
  Frequent self-contradictions or impossible facts; clearly violates core identity information.
\end{itemize}

\paragraph{Dimension 3: Interruption / Backchannel Reasonableness.}
\textbf{Question:} Are interruptions and backchannels used in appropriate places and amounts?
\begin{itemize}
  \item \textbf{9--10: Very reasonable.}
  They occur at natural moments such as long turns, high-information segments, or emotional peaks; they are inserted at appropriate positions and improve conversational rhythm.
  \item \textbf{7--8: Mostly reasonable.}
  Generally appropriate usage, with minor overuse or underuse; a few instances feel slightly off, but overall acceptable.
  \item \textbf{5--6: Fair.}
  The intention to include interruptions and backchannels is clear but somewhat coarse; placement or frequency can feel awkward or repetitive.
  \item \textbf{3--4: Poor.}
  Very noisy usage, with either almost no such events or too many; often inserted where they are not needed or where they disrupt semantics.
  \item \textbf{1--2: Very poor.}
  Usage is almost always unreasonable or violates the intended specification; seriously harms readability.
\end{itemize}

\paragraph{Dimension 4: Human-Likeness.}
\textbf{Question:} Does the speaking style feel human, with natural disfluency, hesitation, and emotion?
\begin{itemize}
  \item \textbf{9--10: Highly human-like.}
  Natural fillers and hesitations are present but not overused; emotions and attitudes are clear; speakers have distinct speaking styles.
  \item \textbf{7--8: Quite human-like.}
  Some colloquial markers are present, such as ``uh'', ``honestly'', or ``to be fair''; emotions are present but not very rich; slight template flavor may remain.
  \item \textbf{5--6: Fair.}
  Occasional fillers or affective words appear, but the style remains somewhat formulaic; many lines still read like information dumps rather than lived speech.
  \item \textbf{3--4: Poor.}
  Either almost no colloquial traces, or mechanical overuse of fillers; very little emotional coloring; monotone style.
  \item \textbf{1--2: Very poor.}
  Reads like formal documentation or rigid templates, or contains strange and unrealistic fillers that feel obviously fake.
\end{itemize}

\paragraph{Rationale Quality Rubric.}
We further ask raters to evaluate rationale annotations on a 1--10 scale along six dimensions.

\paragraph{Dimension 1: Reasonableness / Discourse-Function Accuracy.}
\textbf{Question:} Does the rationale correctly describe the discourse function of this second-level snippet, such as recommendation, explanation, response, shift, example, clarification, or topic transition?
\begin{itemize}
  \item \textbf{9--10:} Clear and accurate function judgment; tightly matches the intent of the current second.
  \item \textbf{7--8:} Mostly accurate, with minor generalization that does not harm correctness.
  \item \textbf{5--6:} Generally reasonable but somewhat vague or off the main point.
  \item \textbf{3--4:} Partly unreasonable or misclassifies the function, such as treating an explanation as a question.
  \item \textbf{1--2:} Largely invalid or opposite to the intended meaning.
\end{itemize}

\paragraph{Dimension 2: Context Grounding.}
\textbf{Question:} Is the rationale grounded in information visible up to the current second, rather than generic commentary?
\begin{itemize}
  \item \textbf{9--10:} Clearly grounded in prior visible context, with correct reference and continuation.
  \item \textbf{7--8:} Context dependence is evident, but the reference is somewhat broad.
  \item \textbf{5--6:} Weak grounding; sounds more like generic dialogue commentary.
  \item \textbf{3--4:} Forced linkage; could apply to almost any dialogue.
  \item \textbf{1--2:} Essentially unrelated to the context, or links to the wrong context.
\end{itemize}

\paragraph{Dimension 3: Intra-Utterance Coherence.}
\textbf{Question:} Does the rationale explain how the current second relates to the earlier part of the same utterance, such as continuation, expansion, contrast, evidence, or sharpening an example?
\begin{itemize}
  \item \textbf{9--10:} Explicitly identifies the intra-utterance relation and strongly aligns with the immediately preceding content.
  \item \textbf{7--8:} Mentions intra-utterance linkage but with less precision.
  \item \textbf{5--6:} Uses vague continuity language, such as ``continues'' or ``adds'', without specifying the relation type.
  \item \textbf{3--4:} The intra-utterance relation is partly incorrect, such as calling an example a contrast.
  \item \textbf{1--2:} No intra-utterance relation is captured, or the relation is clearly wrong.
\end{itemize}

\paragraph{Dimension 4: Inter-Utterance Linkage.}
\textbf{Question:} When cross-utterance linkage is needed, such as responding, returning to a prior topic, comparing two entities, or reactivating a previously mentioned entity, does the rationale capture it accurately? When it is not needed, does it avoid forcing it?
\begin{itemize}
  \item \textbf{9--10:} Links across utterances accurately when appropriate and avoids forced cross-linking when unnecessary.
  \item \textbf{7--8:} Cross-link direction is correct but slightly broad or mildly forced.
  \item \textbf{5--6:} Cross-linking is present but vague, or shows mild over-referencing.
  \item \textbf{3--4:} Clearly forced cross-linking, or links to an unrelated earlier utterance.
  \item \textbf{1--2:} Needed cross-linking is missing entirely, or the cross-link is completely wrong.
\end{itemize}
\noindent
\textit{Key principle:} Cross-utterance linkage is optional but must be accurate; it should not be forced into every second.

\paragraph{Dimension 5: Specificity \& Focus.}
\textbf{Question:} Does the rationale focus on the incremental contribution of the current second, rather than summarizing a larger span?
\begin{itemize}
  \item \textbf{9--10:} Tightly captures what is newly added in the current second, such as a micro-step from recommendation to supporting reason.
  \item \textbf{7--8:} Mostly focused, with minor extra expansion.
  \item \textbf{5--6:} Somewhat summarizes the whole utterance; second-level granularity is insufficient.
  \item \textbf{3--4:} Clearly summarizes a larger span or drifts off-topic.
  \item \textbf{1--2:} Fails to capture the incremental point of the current second.
\end{itemize}

\paragraph{Dimension 6: Clarity \& Non-Template Style.}
\textbf{Question:} Is the rationale natural, clear, and informative, while avoiding repetitive template phrasing across seconds?
\begin{itemize}
  \item \textbf{9--10:} Natural, concise, and specific; does not read like a template.
  \item \textbf{7--8:} Clear overall, with occasional templated phrasing.
  \item \textbf{5--6:} Understandable but repetitive or overly formulaic.
  \item \textbf{3--4:} Strongly templated with low information content.
  \item \textbf{1--2:} Hard to read, ungrammatical, or mostly empty filler.
\end{itemize}

\section{Method Details}

\subsection{Behavior Perceiver Implementation}

\begin{tcolorbox}[
    title={Implementation details of the behavior perceiver},
    colback=gray!3,
    colframe=gray!60,
    fonttitle=\bfseries,
    breakable,
    enhanced,
    boxrule=0.5pt,
    arc=2pt,
    left=4pt,
    right=4pt,
    top=4pt,
    bottom=4pt
]
\begin{lstlisting}[basicstyle=\ttfamily\scriptsize, breaklines=true]
HB, HE in R^{B x T x 768}

Fusion:
  lam = sigmoid(Wb(HB) + We(HE))                  # Wb, We: Linear(768 -> 768)
  e = (1 - lam) * HB + lam * HE                  # R^{B x T x 768}

Shared latent:
  z = Linear(768 -> 768) -> GELU -> Dropout(0.1) -> Linear(768 -> 768)
  z_seq in R^{B x T x 768}

CausalEvidenceTrunk:
  prev_t = z_{t-1}, with prev_0 = z_0
  delta_t = z_t - prev_t

  short = DWConv1d(k=5, groups=768) -> PWConv1d(1x1) -> GELU
  short is computed on left-padded normalized input, so it is causal.

  joint_t = concat[
    LN(z_t), LN(prev_t), LN(delta_t), short_t
  ]                                              # 4 * 768 = 3072 dims

  hidden_t = Linear(3072 -> 768) -> GELU -> Dropout(0.1)
  hidden_seq = GRU(input=768, hidden=768, batch_first=True)
  evidence_t = LN -> GELU -> Dropout(0.1) -> Linear(768 -> 768)

HighHead:
  x_t = LN(evidence_t) -> Linear(768 -> 768) -> GELU -> Dropout(0.1)
  logits_high_t = Linear(768 -> 4)(x_t)
  guidance_t = tanh(Linear(768 -> 192)(x_t))

LowStateBeliefFilter:
  belief dim = 5 labels x 4 duration buckets = 20
  initial belief b_init = one-hot(state index 0)

  stay/switch head:
    s_t = MLP([LN(evidence_t), LN(guidance_t), LN(b_{t-1})])
        = Linear(980 -> 256) -> GELU -> Dropout(0.1) -> Linear(256 -> 2)

  next-label head:
    n_t = MLP([LN(evidence_t), LN(guidance_t), LN(b_{t-1})])
        = Linear(980 -> 256) -> GELU -> Dropout(0.1) -> Linear(256 -> 5)

  p_t = softmax(s_t)                              # [stay, switch]
  q_t = softmax(n_t)                              # next low-label distribution

  duration_shift(b):
    reshape b to [B, 5, 4]
    bucket d moves to d+1
    the last bucket is saturating / absorbing

  reset_from_labels(q):
    reshape to [B, 5, 4]
    put q into duration bucket 0 only

  b_t = p_t[stay] * duration_shift(b_{t-1})
      + p_t[switch] * reset_from_labels(q_t)

LowReadoutHead:
  joint_t = concat[LN(evidence_t), LN(b_t)]       # 768 + 20 = 788 dims
  hidden_t = Linear(788 -> 768) -> GELU -> Dropout(0.1)
           -> Linear(768 -> 768) -> GELU
  logits_low_t = LN -> Dropout(0.1) -> Linear(768 -> 5)
\end{lstlisting}
\end{tcolorbox}

\subsection{Training Details}

\paragraph{Input Processing.}
All audio inputs are resampled to 16~kHz and segmented into 1-second chunks. The ASR stream is updated every 0.5 seconds, while features and labels are aggregated into 1.0-second bins.

\paragraph{Behavior Perceiver Training.}
We train the behavior perceiver for 20 epochs with a batch size of 8 and a random seed of 42. Mixed-precision training is enabled with FP16, using AMP autocast and GradScaler. We use AdamW with a learning rate of $1\times10^{-3}$ for all trainable parameters. A linear warmup followed by linear decay schedule is used. The number of warmup steps is set to $\min(1000, \text{total training steps})$. Since this run contains $20 \times 149 = 2980$ total training steps, the effective warmup length is 1000 steps. Gradient clipping is applied with a maximum gradient norm of 1.0.

The high-level and low-level objectives are defined separately and linearly combined during training. For high-level speech-act prediction, we use class-weighted cross-entropy:
\[
\mathcal{L}_{\mathrm{high}}
=
\mathrm{CE}(\mathbf{z}^{h}, y^{h}; \mathbf{w}^{h}),
\]
where positions with label $-100$ are ignored. The class weights are computed from the training-set class frequencies as:
\[
w^{h}_{c}
=
\frac{N_{h}}{C_{h}\max(n^{h}_{c}, 1)},
\]
where $N_{h}$ is the total number of valid high-level labels, $C_{h}$ is the number of high-level classes, and $n^{h}_{c}$ is the number of training labels in class $c$.

For low-level interaction-behavior prediction, we use class-weighted focal loss rather than standard cross-entropy. The focal parameter is set to $\gamma=2.0$. The low-level class weights are computed analogously:
\[
w^{l}_{c}
=
\frac{N_{l}}{C_{l}\max(n^{l}_{c}, 1)}.
\]
For each valid low-level label position, the loss is:
\[
\mathcal{L}_{\mathrm{low}}
=
- w^{l}_{y}(1-p_{t})^{\gamma}\log(p_{t}),
\]
where $p_{t} = \mathrm{softmax}(\mathbf{z}^{l})_{y}$. The final low-level loss is averaged over all valid positions in the batch, and labels with value $-100$ are ignored.

The total training objective is:
\[
\mathcal{L}
=
\lambda_{\mathrm{high}}\mathcal{L}_{\mathrm{high}}
+
\lambda_{\mathrm{low}}\mathcal{L}_{\mathrm{low}}.
\]
We set $\lambda_{\mathrm{high}}=\lambda_{\mathrm{low}}=1.3$, so the two losses are weighted symmetrically in this run.

Although the model computes stay/switch logits and next-label logits in the forward pass, these outputs do not receive separate auxiliary supervision. During training, only \texttt{logits\_high} and \texttt{logits\_low} are directly supervised. Consequently, the belief filter, persistence head, and next-state proposal head are learned indirectly through the final low-level focal loss, without additional transition-state BCE or CE auxiliary losses.

\paragraph{GoT Training Objective.}
The GoT module is trained with two objectives: a Stage-1 selector loss for retrieving causal evidence sentences and a Stage-2 decoder loss for generating per-second rationales. We set the selector temperature to $T=1.0$. Sentence candidates are formed from the past window $[t-W,t)$ with $W=90$ seconds.

For each sample $(\mathrm{audio\_id}, t)$, the selector outputs candidate sentence scores $s_{t,j}$ and a dynamic threshold $\tau_t$. We define the threshold-aligned logits as:
\[
\ell_{t,j}
=
\frac{s_{t,j}-\tau_t}{T}.
\]
Given multi-hot labels $y_{t,j}\in\{0,1\}$ and a candidate mask $m_{t,j}$, the Stage-1 selector objective is:
\[
\mathcal{L}_{\mathrm{sel}}
=
\mathcal{L}_{\mathrm{wbce}}
+
\lambda_{\mathrm{count}}\mathcal{L}_{\mathrm{count}}
+
\lambda_{\mathrm{rank}}\mathcal{L}_{\mathrm{rank}}.
\]
We use weighted BCE with logits, where the positive-class weight is computed once on the training split and kept fixed as $\alpha = N_{\mathrm{neg}}/N_{\mathrm{pos}}$ over candidate positions. The weighted binary cross-entropy term is computed only over valid candidates:
\[
\mathcal{L}_{\mathrm{wbce}}
=
\frac{1}{\sum_j m_{t,j}}
\sum_j
m_{t,j}\,
\mathrm{WBCE}(\ell_{t,j}, y_{t,j}; \alpha).
\]
We set $\lambda_{\mathrm{count}}=0.01$ and $\lambda_{\mathrm{rank}}=0.1$. The count regularizer is:
\[
\mathcal{L}_{\mathrm{count}}
=
\left(
\sum_j \sigma(\ell_{t,j}) - \sum_j y_{t,j}
\right)^2,
\]
which encourages the predicted number of selected evidence sentences to match the number of positive labels. To promote positive-sample ranking, we use:
\[
\mathcal{L}_{\mathrm{rank}}
=
\frac{1}{|P_t|}
\sum_{j\in P_t}
-\log
\frac{\exp(\ell_{t,j})}{\sum_k \exp(\ell_{t,k})},
\]
where $P_t=\{j:m_{t,j}=1, y_{t,j}=1\}$ is the set of valid positive candidates.

For the Stage-2 T5 rationale decoder, given the tokenized input $\mathbf{x}_t$ formed by the linearized causal evidence chain and the target rationale tokens $\mathbf{y}_t=(y_{t,1},\ldots,y_{t,L})$, we use standard teacher-forced seq2seq negative log-likelihood:
\[
\mathcal{L}_{\mathrm{dec}}
=
-\sum_{n=1}^{L}
\log p_{\theta}(y_{t,n}\mid y_{t,<n}, \mathbf{x}_t).
\]
The selector loss supervises causal evidence retrieval, while the decoder loss supervises per-second rationale generation conditioned on the selected evidence chain.

\subsection{Strictly Causal Protocol}

Our strictly causal convention governs the entire pipeline, including data construction, annotation generation, model training, and deployment. During ConversationGoT-120h construction, the annotation system is not allowed to access any information beyond the current time step, and it is also prohibited from exploiting implicit future cues, such as how the current utterance will continue in later seconds. Correspondingly, when generating speech-act and rationale annotations, GPT-5 is constrained to use only the current second and the preceding history as input. This prevents causal leakage where labels are filled in after observing the future and ensures that annotations are producible from online streaming observations.

The same constraint is applied to OOD annotation on the Candor dataset. We use a strictly causal streaming ASR system that outputs transcripts only up to the current time, and all subsequent modules, including GPT-5 and any auxiliary components, are forbidden from accessing future audio, future transcripts, or future semantic cues. This makes the OOD annotation process consistent with a real online perception setting. Aligned with these data- and annotation-side constraints, all proposed modules also operate under a strictly causal setting during both training and deployment: at every time step, predictions depend only on the current input and historical states, never on future segments.

\section{Experiment}
\subsection{Evaluation Metrics}

For the conversational behavior state perceiver, we report F1 and AUC for both high-level communicative-function prediction and low-level interaction-behavior prediction. F1 is the harmonic mean of precision and recall:

\[
\mathrm{F1}
=
\frac{2\cdot \mathrm{Precision}\cdot \mathrm{Recall}}
{\mathrm{Precision}+\mathrm{Recall}}.
\]

A higher F1 indicates fewer false positives and false negatives at the selected decision threshold.

AUC is the area under the receiver operating characteristic curve. It measures the model's threshold-independent ability to discriminate positive examples from negative examples. A higher AUC indicates stronger overall ranking and discrimination performance.

\paragraph{Human--Model Agreement Protocol.}
For each evaluated second, raters are shown the model-predicted high-level communicative function and low-level interaction behavior and asked whether they agree with each label. Human--Model Agreement (HMA) is computed as the average binary agreement over evaluated seconds and raters. We report HMA separately for high-level functions, denoted HMA$^{h}$, and low-level interaction behaviors, denoted HMA$^{l}$.

\subsection{Baseline Label Alignment}
\label{app:baseline_label_alignment}

Because external baselines use label spaces that differ from our conversational behavior taxonomy, we define task-specific alignment rules before evaluation.

\paragraph{MIDAS to High-level Communicative Functions.}
For the high-level communicative-function baseline, we use MIDAS and explicitly map its original 23 dialog-act labels to our four high-level communicative-function categories. The original MIDAS label inventory is taken from \texttt{midas\_classifier.py}, and the evaluation-time mapping is implemented in \texttt{midas\_highsa\_eval.py}. The mapping is as follows:

\begin{table*}[h]
\centering
\caption{Mapping from MIDAS dialog-act labels to our high-level communicative-function categories.}
\label{tab:midas_highsa_mapping}
\small
\begin{tabular}{ll}
\toprule
\textbf{High-level category} & \textbf{MIDAS labels} \\
\midrule
Constatives &
\begin{tabular}[t]{@{}l@{}}
statement, opinion, pos\_answer, neg\_answer, other\_answers, \\
comment, complaint, abandon, nonsense, other
\end{tabular}
\\[0.6em]

Directives &
\begin{tabular}[t]{@{}l@{}}
command, dev\_command, open\_question\_factual, \\
open\_question\_opinion, yes\_no\_question
\end{tabular}
\\[0.6em]

Acknowledgments &
\begin{tabular}[t]{@{}l@{}}
back-channeling, appreciation, thanking, hold, apology, \\
respond\_to\_apology, opening, closing
\end{tabular}
\\[0.6em]

Commissives &
$\emptyset$
\\
\bottomrule
\end{tabular}
\end{table*}

Under this mapping, MIDAS does not provide any original dialog-act label that is mapped to our Commissives category. We therefore leave Commissives unmapped for the MIDAS baseline and report the corresponding score as not applicable.

\paragraph{TalkingTurns to Low-level Interaction Behaviors.}
For the low-level interaction-behavior baseline, we use the supervised turn-taking judge from TalkingTurns. Its event labels are naturally aligned with our low-level interaction-behavior taxonomy, including turn-taking, interruption, backchannel, continuation, and silence-like non-event regions. Therefore, no additional many-to-one label mapping is required for this baseline. We directly evaluate the predicted low-level interaction behaviors after temporal alignment to our one-second evaluation grid.

\paragraph{Runtime Analysis.}
We measure latency under a streaming inference setting with batch size 1 on a single NVIDIA RTX A6000 GPU. The system processes one-second audio chunks online and reports input-to-rationale latency, defined as the time from the arrival of the current audio chunk to the completion of rationale decoding. As shown in Table~\ref{tab:latency_breakdown}, S-MARC achieves an average input-to-rationale latency of 0.8249,s per one-second tick. Its steady-state average latency is 0.8361,s, and its p95 latency is 1.0491,s. The component-level breakdown shows that streaming ASR update and D0 prefix decoding dominate the online computational cost, while text encoding and fusion introduce negligible overhead. These results indicate that S-MARC operates within the one-second streaming budget on average. Although the p95 latency slightly exceeds one second, the average latency still leaves a margin of approximately 0.175,s, suggesting that the system offers good real-time usability in standard streaming scenarios.

\begin{table}[t]
\centering
\small
\begin{tabular}{lc}
\hline
\textbf{Component} & \textbf{Mean latency} \\
\hline
Streaming ASR update & 91.9\,ms \\
Frozen acoustic encoder & 32.3\,ms \\
Text encoder & 7.7\,ms \\
Fusion + shared latent projection & 0.3\,ms \\
D0 prefix decoding & 44.5\,ms \\
\hline
End-to-end input-to-rationale latency & 0.8249\,s \\
Steady-state latency & 0.8361\,s \\
p95 latency & 1.0491\,s \\
\hline
\end{tabular}
\caption{Latency breakdown under the streaming inference setting. All measurements are conducted with batch size 1 on a single NVIDIA RTX A6000 GPU. The system processes one-second audio chunks online, and end-to-end latency is measured from the arrival of the current audio chunk to the completion of rationale decoding.}
\label{tab:latency_breakdown}
\end{table}

\paragraph{Real-conversation robustness.}

To further examine whether S-MARC generalizes beyond the mixed test set, we separately evaluate low-level interaction-behavior prediction on the Candor-only subset. Table 11 reports per-class F1 scores on the synthetic-only test set and the Candor-only subset. Overall, S-MARC achieves a macro-F1 of 0.535 on the mixed test set and 0.578 on the Candor-only subset, suggesting that the low-level behavior detector remains effective on naturalistic real conversations. In particular, the Candor-only subset yields higher F1 scores for Silence, Interruption, Backchannel, and Continuation, with Backchannel improving from 0.361 to 0.551 and Interruption improving from 0.287 to 0.382. These results indicate that the model can capture several fine-grained interaction behaviors in real conversational data, rather than relying solely on regularities from synthetic dialogues.

At the same time, the Candor-only result also reveals a remaining limitation. Turn-taking F1 drops from 0.624 on the mixed test set to 0.408 on the Candor-only subset. This suggests that smooth speaker transitions in natural conversations are harder to detect than more explicit interaction events such as silence, backchannels, or interruptions. Real conversations often contain gradual floor transfer, ambiguous overlap, hesitation, and short listener responses, which can blur the boundary between Turn-taking, Backchannel, and Continuation. Therefore, while the Candor-only evaluation provides preliminary evidence of robustness to real dialogue, it also highlights the need for stronger modeling of turn-transition boundaries under natural conversational timing.

\begin{table}[t]

\centering

\small

\begin{tabular}{lcc}

\hline

\textbf{Low-level behavior} & \textbf{synthetic-only} & \textbf{Candor-only} \\

\hline

Silence      & 0.594 & 0.708 \\

Turn-taking  & 0.624 & 0.408 \\

Interruption & 0.287 & 0.382 \\

Backchannel  & 0.361 & 0.551 \\

Continuation & 0.809 & 0.838 \\

\hline

Macro-F1     & 0.535 & 0.578 \\

\hline

\end{tabular}

\caption{Low-level interaction-behavior per-class F1 scores on the synthetic-only test set and the Candor-only subset. The Candor-only evaluation assesses robustness on naturalistic real conversations.}

\label{tab:low_level_candor_f1}

\end{table}

\end{document}